\documentclass[final,12pt]{clear2025} 

\newtheorem{assumption}{Assumption}

\newtheorem*{lemE*}{Lemma}

\newtheorem*{proE*}{Property}

\usepackage[createShortEnv]{proof-at-the-end}

\usepackage{tcolorbox}
\usepackage{enumitem}
\usepackage{algorithm,algpseudocode}
\usepackage{booktabs}
\usepackage{colortbl}
\usepackage{pgfplots}
\pgfplotsset{compat=newest}
\pgfplotsset{scaled y ticks=false}
\usepgfplotslibrary{groupplots}
\usepgfplotslibrary{dateplot}
\usepackage{tikz}
\pgfplotsset{compat=1.11,
 /pgfplots/ybar legend/.style={
 /pgfplots/legend image code/.code={
 \draw[##1,/tikz/.cd,yshift=-0.25em]
 (0cm,0cm) rectangle (3pt,0.8em);},
 },
}
\newcommand\independent{\protect\mathpalette{\protect\independenT}{\perp}}
\def\independenT#1#2{\mathrel{\rlap{$#1#2$}\mkern2mu{#1#2}}}

\hypersetup{
 colorlinks=true,
 linkcolor=cyan,
 filecolor=magenta, 
 urlcolor=cyan,
 pdftitle={Overleaf Example},
 pdfpagemode=FullScreen,
 citecolor	= magenta,
 }

\title[On Measuring ICC in DNN]{On Measuring Intrinsic Causal Attributions in Deep Neural Networks}
\usepackage{times}

 \clearauthor{\Name{Saptarshi Saha}\thanks{Saptarshi was a Fulbright-Nehru Doctoral Research Fellow at the University at Buffalo during  this work.}\Email{saptarshi.saha\_r@isical.ac.in}\\
 \Name{Dhruv Vansraj Rathore} \Email{cs2306@isical.ac.in}\\
 \Name{Soumadeep Saha}\Email{soumadeep.saha\_r@isical.ac.in}\\
 \Name{Utpal Garain} \Email{utpal@isical.ac.in}\\
 \addr Indian Statistical Institute, Kolkata, West Bengal - 700108, India
 \AND
 \Name{David Doermann} \Email{doermann@buffalo.edu}\\
 \addr University at Buffalo, Buffalo, NY, USA
 }

\begin{document}

\maketitle

\begin{abstract}%
Quantifying the causal influence of input features within neural networks has become a topic of increasing interest. Existing approaches typically assess direct, indirect, and total causal effects. This work treats NNs as structural causal models (SCMs) and extends our focus to include intrinsic causal contributions (ICC). We propose an identifiable generative post-hoc framework for quantifying ICC. We also draw a relationship between ICC and Sobol' indices.
Our experiments on synthetic and real-world datasets demonstrate that ICC generates more intuitive and reliable explanations compared to existing global explanation techniques.
\end{abstract}


\section{Introduction}
In recent years, there has been a significant surge of interest in incorporating causal principles into deep learning models \citep{pawlowski2020dscm, saha2022on}. Much of the existing research has focused on post-hoc explanations of trained neural networks' decisions using causal effect analysis \citep{pmlr-v97-chattopadhyay19a,alvarez-melis-jaakkola-2017-causal}. Other studies have explored counterfactuals for explanations or data augmentation \citep{dash2022evaluating,pmlr-v97-goyal19a,reddy2023counterfactualdataaugmentationconfounding, NEURIPS2020_294e09f2}, causal disentangled representation learning \citep{9578520,9363924,JMLR:v23:21-0080}, and causal discovery methods \citep{Zhu2020Causal}. However, despite efforts \citep{pmlr-v97-chattopadhyay19a,Reddy2023TowardsLA,Kancheti2021MatchingLC} to quantify the causal attributions learned by neural networks, there is presently no viable method for elucidating the ``intrinsic causal contribution" (ICC) \citep{janzing2024quantifyingintrinsiccausalcontributions} in neural networks. In this paper, we present a new framework based on generative models—the first of its kind, to the best of our knowledge—that quantifies intrinsic causal contributions in neural network models. 
To illustrate this concept of ICC, imagine a relay race with three runners: $A$, $B$, and $C$. Runner $A$ starts the race and passes the baton late to runner $B$, who then hands it off late to runner $C$, who ends up finishing late as well. To determine the ``intrinsic contribution'' of runner B to the delay of runner $C$, we compare the delay of $C$ to a situation where $B$ only contributes the delay it inherited from $A$ without adding any additional delay of its own. This means we're looking at how much delay $B$ is responsible for beyond what it received from $A$. This concept helps differentiate between delays that $B$ causes itself (intrinsic) and delays it simply passes on from $A$ (inherited).
This distinction is meaningful whether we analyze the delay in a single race, the average delay across many races, or the variation in delays across multiple races. 


To motivate the need for studying intrinsic causal contributions in neural network models, let's consider the task of predicting a patient's recovery time ($R$) using the features: treatment type ($T$), initial health condition ($H$), and post-treatment care ($P$). In the real world, $H$ influences both $T$ and $P$; while $T$ also influences $P$. $H$, $T$, and $P$ all influence $R$. However, these relationships among the input features $H$, $T$, and $P$ are often not explicitly modeled in a neural network model. Now, assume that patients with severe initial health conditions are assigned to more aggressive treatment. It is possible that a neural network model might misattribute the longer recovery times directly to aggressive treatments without considering the severity of the initial health condition. Usual causal effect estimates the expected change in Recovery Time $R$ as the treatment $T$ changes. It doesn’t account for the effect of upstream variable $H$ on $T$ (due to the do-intervention on $T$). With intrinsic attribution analysis, the model aims to understand the part of $T$’s impact on $R$ that is inherited from $H$, and the part that represents $T$’s intrinsic effect.
Thus, learning intrinsic causal attributions can also find
application in medicine. For example, medical practitioners can look at treatments that have shown intrinsic benefits and consider optimizing these treatments for broader patient use. 

To this end, the aim of our work is to identify the intrinsic causal contribution of an input on the output of a neural network.
Our main contributions can be summarized as follows: We introduce an identifiable framework for computing intrinsic causal attributions in neural networks, a concept previously unexplored in neural network attribution to our knowledge. In addition to Shapley-based contributions, we advocate for asymmetric ICC. In Section \ref{Axioms}, we demonstrate that ICC meets several desirable properties for an attribution method. In Section \ref{sobolsection}, we establish connections between the ICC and Sobol indices, offering a fresh perspective on global sensitivity analysis from a causal viewpoint. Finally, our experiments show that the ICC produces reliable global explanations.

\section{Related works}

\paragraph{ Explainability} Several established methods for explaining neural network models quantify the influence of input features on model outputs. 
These methods include saliency
maps \citep{deepinsidecnn,Deconvolutional,GradCam}, Locally Interpretable
Model-Agnostic Explanations (LIME) \citep{LIME}, Integrated Gradients \citep{IG}, DeepLift \citep{DeepLIFT}, Shapley values \citep{SHAP} among others. While some of these techniques are model-agnostic, they are local in nature, meaning that the explanations are limited to individual predictions. 
On the other hand, global attributions are a powerful tool for interpretability because they highlight the importance of features across an entire population. They often use interpretable surrogate models like decision trees or adjust the input space to assess overall predictive power \citep{Lakkaraju2016InterpretableDS, Frosst2017DistillingAN, 8622994}. Submodule pick LIME
(SP-LIME) \citep{SP_LIME} uses submodular optimization to summarize local attributions, better capturing learned interactions. However, like surrogate models, it extracts useful and independent explanations from the LIME method, which may not effectively capture the non-linear feature interactions learned by neural networks. \citet{GAM} proposed Global Attribution Mapping (GAM) to explain the non-linear representations learned by a neural network across different subpopulations. GAM clusters similar local feature importances to create human-interpretable global attributions, each tailored to explain a specific subpopulation. Additionally, GAM allows for adjustable granularity to capture varying numbers of subpopulations in its global explanations. Permutation Feature Importance (PFI) \citep{Breiman2001-mn,Strobl2008-ey} is another comparable measure across model types, offering a global view of the model's reliance on each feature. However, none of these methods account for causality in their explanations.

\paragraph{Causal Explanations} \cite{frye2021shapley} proposed Asymmetric Shapley Values to integrate real-world causal knowledge by restricting feature permutations to those that align with a (partial) causal ordering. 
\cite{Causal_Shapley_value} introduced causal Shapley values that account for the causal relationships among features to explain their total causal effect on predictions. \cite{do-shap} presented the
do-Shapley values to measure the strengths of different causes to a target quantity.
\cite{pmlr-v97-chattopadhyay19a} proposed a post-hoc explanation method to find average causal effects in a trained neural network by treating it as an structural causal model (SCM). It prompts further studies \citep{9506118,9982682,10.5555/3454287.3455204,DBLP:journals/corr/abs-1907-07165} to quantify learned causal effects more comprehensively. \cite{Reddy2023TowardsLA}
introduce an ante-hoc method that identifies and retains direct, indirect, and total causal effects during the neural network model training process. Other causal explanation methods \citep{verma2022counterfactualexplanationsalgorithmicrecourses,pmlr-v97-goyal19a, wachter2018counterfactualexplanationsopeningblack, 10.1007/978-3-030-58112-1_31,10.1145/3461702.3462597, Mahajan2019PreservingCC, 10.1007/978-3-030-86520-7_40} leverage counterfactuals to examine model behavior under semantically meaningful input changes. \cite{CAGE} propose a causality-aware, model-agnostic framework based on Shapley values for global explanations. However, none of the existing work attempts to quantify ICC for attributions in deep neural networks.

\paragraph{Sensitivity Analysis} Sensitivity analysis (SA) \citep{Saltelli2008GlobalSA} studies how model inputs influence outputs and is widely used to explain input-output relationships in complex systems.
\citet{scholbeck2024positionpaperbridginggap} argue that interpretable machine learning is essentially a form of sensitivity analysis applied to machine learning models. \cite{Look_at_variance} used Sobol' indices to model the attributions of image regions. \cite{Kuhnt2022, 9903639} and \cite{ scholbeck2024positionpaperbridginggap} present an overview of sensitivity analysis methods for interpreting ML models. \cite{TUNKIEL2020107630} apply derivative-based sensitivity analysis to rank high-dimensional features in a directional drilling model. \cite{9903639} use the Morris method to calculate sensitivity indices for genomic prediction. \cite{Benesse2024-sx} demonstrate how fairness can be defined within a global sensitivity analysis (GSA) framework, highlighting shared indicators between the two fields. They also demonstrate how GSA frameworks can address causal fairness, using specific Sobol' indices to detect causal links between sensitive variables and algorithm outcomes. The generalization of Sobol indices within a causal framework remains largely unexplored.

\section{NNs through the lens of SCMs}

\paragraph{Notation} Each random variable is denoted by an uppercase letter (e.g., $V$) and its realized value by the corresponding lowercase letter (e.g., 
$v$). We use boldface letters 
$\mathbf{V}$ and 
$\mathbf{v}$ to represent a set of variables and their corresponding realized values, respectively. The set $\{1, \ldots, p\}$ is denoted as $[p]$. As we often need to work with $A \cup \{j\}$, it is handy to write $A+j$ for it. $A-j$ represents the set difference $A\setminus \{ j\}$. Throughout this work, we use $P$ to denote probability distributions and 
$\tilde{p}$ to represent the corresponding density or probability mass functions (e.g., $P(X_{1})$ vs. $\tilde{p}(x_{1})$). 

\begin{figure}[!htb] 
 \centering 
 \includegraphics[width=10cm,]{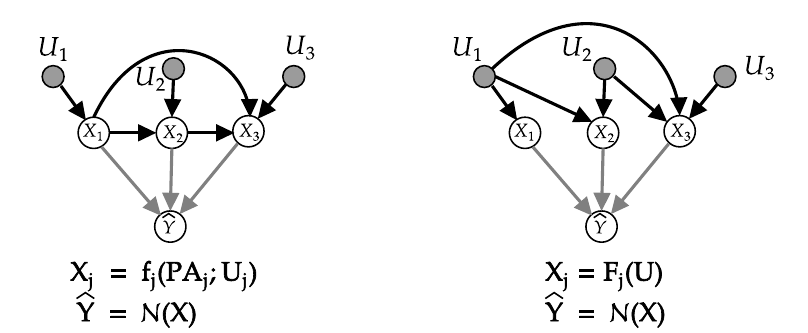}
 \caption{ An example of a causal view of a NN with three input features. White nodes represent variables that are either observed or assumed to be known, while shaded nodes indicate unobserved or latent variables. 
 The left graph illustrates the causal relationships between features along with their exogenous parents, while the right graph utilizes exogenous variables for the TMI mapping of the SCM of inputs. In both figures, the grey edges serve to augment the neural network to the SCM. 
 }
 \label{scm_view_of_nn} 
\end{figure}

\paragraph{}
This work is grounded in the principles of causality, specifically SCMs and the do-calculus, as outlined by \citet{Pearl_2009}. A concise overview of the relevant concepts is provided in the Appendix \ref{SCM}.
Consider a causal graph $\mathcal{G}=(\mathbf{X},\mathcal{E})$, where $\mathbf{X}=\{X_{1},X_{2},...,X_{p} \}$ represents the set of input features (random variables), and 
$\mathcal{E}$ denotes the set of edges that capture the causal relationships among the variables in 
$\mathbf{X}$.
\begin{assumption}\label{assumption1}
The causal graph
$\mathcal{G}$ is acyclic and contains no latent (unobserved) confounders.
\end{assumption}
Let $\mathcal{N}$ be a neural network model that has been trained to predict $Y$ from input features $\mathbf{X} $ by minimizing the empirical loss.
The neural network 
$\mathcal{N}$ can be envisioned as a directed acyclic graph (DAG) consisting of directed edges that link successive layers of neurons. Consequently, the predicted output $\hat{Y}=\mathcal{N}(\mathbf{X})$
can be interpreted as the outcome of a sequence of interactions from the initial layer to the final layer of the network $\mathcal{N}$. When analyzing the intrinsic contributions of inputs on the output of $\mathcal{N}$, only the neurons in the first and final layers are considered. Therefore, akin to the approach in \cite{pmlr-v97-chattopadhyay19a, Kancheti2021MatchingLC} we can marginalize the influence of the hidden layers within 
$\mathcal{N}$ and concentrate exclusively on the causal structure between inputs and outputs. 
With our view of a neural network as an SCM, we define augmented causal graph $\Tilde{\mathcal{G}} = (\mathcal{V},\Tilde{\mathcal{E}})$ with $\mathcal{V}= \mathbf{X} \cup\{\hat{Y} \}$ and $\Tilde{\mathcal{E}}=\mathcal{E}\cup \bigcup_{j=1}^{p}\{(X_{j},\hat{Y})\}$.
Note that while our perspective on neural networks as SCMs is the same as \cite{pmlr-v97-chattopadhyay19a, Kancheti2021MatchingLC}, they do not address or model intrinsic causal attribution, which is central to our study. 
To measure the intrinsic contribution of each feature to 
$\hat{Y}$, we first recursively substitute structural equations into one another, expressing each feature $X_{j}$ solely in terms of the unobserved noise variables $\mathbf{U}$:
\begin{align}\label{Stractural_assignment}
 X_{j} = f_{j}(PA_{j};U_{j})=F_{j} (\mathbf{ U}) = F_{j}(U_{1},...,U_{p}), \qquad \forall 1\leq j \leq p. 
\end{align}
Figure \ref{scm_view_of_nn} portrays an example of our SCM perspective on neural networks.
As $\mathcal{G}$ is acyclic, $\mathbf{F} = (F_{1},F_{2}...,F_{p})$ is a triangular map. 
More importantly, any SCM can be represented as a tuple $(\mathbf{F},P_{\mathbf{U}})\in \mathcal{F}\times \mathcal{P_{U}}$, where 
$\mathcal{F}$ denotes the set of all triangular monotonic increasing (TMI) maps, and $\mathcal{P_{U}}$ represents the set of all fully-factorized distributions
$P_{\mathbf{U}}(\mathbf{u})=\prod_{j=1}^{p} P_{U_{j}}(u_{j})$. TMI maps are autoregressive functions where the 
$i$-th component is strictly monotonically increasing with respect to its 
$i$-th input. 
Mathematically, a TMI map is characterized as a function $ T: \mathbb{R}^p \rightarrow \mathbb{R}^p $ defined as follows:
\begin{align*}
T(u) = \begin{bmatrix} T_1(u_1) & T_2(u_1, u_2) & \cdots & T_d(u_1, \ldots, u_p) \end{bmatrix}^\top,
\end{align*}
where 
each component function $ T_k $ depends solely on the first $ k $ variables $ u_{\leq k} := (u_1, \ldots, u_k) $ and is monotone increasing with respect to the last input $ u_k $ for any $ (u_{k+1}, \ldots, u_p) $.
They can approximate any fully supported distribution and can be parameterized through deep neural networks. For more details on TMI maps, please see \cite{DBLP:conf/aistats/XiB23}, \cite{ irons2021triangularflowsgenerativemodeling}. 

\section{Intrinsic Causal Contributions}

Now that we can conceptualize each $X_{j}$ as being influenced by the independent causal factors $U_{1},...,U_{p}$, the change in uncertainty within $\hat{Y}$ is assessed as a result of a hypothetical adjustment of $U_{j}$ (which is standard conditioning stemming from exogeneity). We initiate our discussion by defining the intrinsic causal contributions \citep{janzing2024quantifyingintrinsiccausalcontributions} of input features to the output of a trained neural network model $\mathcal{N}$.

\begin{definition}[Intrinsic Causal Contribution  in Neural Network]\label{def_ICC}

Given an adjustment set $I \subseteq [p]-\{j\}$, the intrinsic causal contribution of a feature $X_{j}$ on the output $\hat{Y}$ of a NN $\mathcal{N}$ is defined as
 \begin{align}\label{ICC}
 ICC_{\phi}(X_{j}\rightarrow \hat{Y}|I ) = \phi(\hat{Y}|\mathbf{U}_{ j + I}) - \phi(\hat{Y}|\mathbf{U}_{I}),
 \end{align}
where $\phi$ is a measure of conditional uncertainty that satisfies one of  the following two conditions: 
\begin{enumerate}[noitemsep,topsep=0pt]
    \item \textbf{monotonicity:} $\phi(\hat{Y}|\mathbf{U}_{I}) \geq \phi(\hat{Y}|\mathbf{U}_{j + I})$, \quad or \quad $\phi(\hat{Y}|\mathbf{U}_{I}) \leq \phi(\hat{Y}|\mathbf{U}_{j + I})$
     \item  \textbf{calibration:}  $\phi(\hat{Y})= \phi(\hat{Y}|\mathbf{U}_{\varnothing})  = 0$, \quad or \quad $\phi(\hat{Y}|\mathbf{U}) = 0$
\end{enumerate}
where $\varnothing$ is the empty set.
In this context, $\phi(\cdot|U_{I})$ denotes conditioning on all noise variables $U_{i}$ for which $i$ is an element of the set $I$.
\end{definition} 

It is important to note that $\hat{Y}$ is a deterministic function of 
$\mathbf{X}$, which differs very slightly from \cite{janzing2024quantifyingintrinsiccausalcontributions}. We adapt the definition in the context of explaining the decisions made by a neural network.
Monotonicity is not an absolute necessity, but it is often simpler to understand and work with positive contributions in practical scenarios. For example, the variance of the conditional expectation can be seen as a choice of $\phi$. We will discuss this in more detail later.
The definition of ICC can be generalized to scenarios involving confounding variables. However, we will not discuss this here to maintain focus on the main discussion. Please look at the foundational ICC paper \cite{janzing2024quantifyingintrinsiccausalcontributions} for more details.

\subsection{Measuring ICC using Shapley value and topological ordering}

Unfortunately, the contribution of each feature $X_{j}$ in (\ref{ICC}) depends on the adjustment set $I$ given as context. \cite{janzing2024quantifyingintrinsiccausalcontributions} address this issue by using Shapley values to symmetrize the ICC: 
\begin{align}\label{ICC_shap}
 ICC_{\phi}^{\text{Sh}}(X_{j}\rightarrow \hat{Y})= \sum_{T \subseteq [p]- j } \dfrac{1}{p \binom{p-1}{|T|} } ICC_{\phi}(X_{j}\rightarrow \hat{Y}| T). 
\end{align}
However, computing the Shapley values can be computationally expensive. Another viable option would be to utilize topological ordering. A topological ordering of a DAG $\mathcal{G}$ is a specific arrangement of the 
its $d$ nodes such that each node is positioned earlier in the sequence than any of its descendants.
Let $S_{p}$ be the symmetric group on the set $[p]$. 
Given an ordering $\pi \in S_{p}$, let $T_{\pi}^{j}$ be the set of indices that occur before $j$ in the ordering $\pi$, i.e., $T_{\pi}^{j} =\{k:\pi(k)<\pi(j) \}$. 
If we define 
\begin{align}\label{ICC_topo}
 ICC_{\phi}^{\pi}(X_{j}\rightarrow \hat{Y})=ICC_{\phi}(X_{j}\rightarrow \hat{Y}| T_{\pi}^{j} ), 
\end{align}
 it is easy to see that
the reliance on arbitrary 
$\pi$ introduces an unnecessary level of ambiguity. Therefore, we may constrain $\pi$ as a topogical (or causal) ordering of the DAG $\mathcal{G}$. However, several valid sequences can meet this requirement, making the ordering non-unique. And
as $\pi$ is not unique, the ambiguity remains unresolved.
Therefore, we adjust (\ref{ICC_topo}) by averaging exclusively over all potential topological orderings of the DAG $\mathcal{G}$ :
\begin{align}\label{ICC_topological}
 ICC_{\phi}^{\text{To}}(X_{j}\rightarrow\hat{Y}) = \dfrac{1}{\lvert \mathcal{C}(\mathcal{G}) \rvert}\sum_{\pi\in \mathcal{C}(\mathcal{G})} ICC_{\phi}(X_{j}\rightarrow\hat{Y}|T_{\pi}^{j}),
\end{align}
where $\mathcal{C}(\mathcal{G})$ is set of all possible causal ordering of $\mathcal{G}$. 
An iterative adaptation of the algorithm proposed by \citet{KNUTH1974153} can be used to generate all possible topological orderings.
The Shapley-based ICC can also be expressed in an alternative, equivalent form \citep{10.5555/3586589.3586632}:
\begin{align}\label{ICC_shapley}
 ICC_{\phi}^{\text{Sh}}(X_{j}\rightarrow\hat{Y}) = \dfrac{1}{p!}\sum_{\pi\in S_{p}} ICC_{\phi}(X_{j}\rightarrow\hat{Y}|T_{\pi}^{j}).
\end{align}
As $|\mathcal{C}(\mathcal{G})|\leq |S_{p}|$, it is clear that from equations \ref{ICC_topological} and \ref{ICC_shapley} that $ICC^{\text{To}}_{\phi}$ is computationally more efficient than $ICC^{\text{Sh}}_{\phi}$. 
We define the value of the coalition for any $T\subseteq[p]$ as $\phi(\hat{Y}|do(\mathbf{X}_{T})):=\sum_{\mathbf{x}_{T^{j}_{\pi}}} \phi\big(\hat{Y}|do(\mathbf{X}_{T}= \mathbf{x}_{T})\big) \tilde{p}(\mathbf{x}_{T})$.

\begin{theoremEnd}[end, restate, text link=, text proof= ]{lemE}[\cite{janzing2024quantifyingintrinsiccausalcontributions}]\label{Identifiability}
For a topological ordering $\pi \in \mathcal{C}(\mathcal{G})$, we have
\begin{align*}
 \phi(\hat{Y}|\mathbf{U}_{T^{j}_{\pi}}) = \phi(\hat{Y}|\mathbf{X}_{T^{j}_{\pi}}) = \phi(\hat{Y}|do(\mathbf{X}_{T^{j}_{\pi}})).
\end{align*}
\end{theoremEnd}

\begin{proofE}
The first equality holds due to the conditional independence $\hat{Y}\independent \mathbf{X}_{T^{j}_{\pi}}|\mathbf{U}_{T^{j}_{\pi}}$ and the fact that $\mathbf{X}_{T^{j}_{\pi}}$ is function of $\mathbf{U}_{T^{j}_{\pi}}$. The second equality is valid as conditioning on all ancestors blocks any backdoor paths.
\end{proofE}

Instead of criticizing $ICC^{\text{To}}_{\phi}$ for avoiding rung 3 causal models \citep[Section~5]{janzing2024quantifyingintrinsiccausalcontributions}, we recognize that Lemma \ref{Identifiability} is crucial for establishing the identifiability of the $ICC^{\text{To}}_{\phi}$.
This means we can measure the causal contribution ($ICC^{\text{To}}_{\phi}$) solely using observational conditionals that are readily estimable from observational data. We do not see $ICC_{\phi}^{\text{To}}$ as a replacement for $ICC_{\phi}^{\text{Sh}}$. The intrinsic value of the latter is undeniable. Instead, we view both formulations as suited for different contexts. Due to space constraints, we provide the causal interpretation of ICC in the Appendix \ref{causal_interpretation}. We also recommend that readers refer to \cite{janzing2024quantifyingintrinsiccausalcontributions} for further insights. Hereafter, when we refer to ICC, it encompasses both $ICC^{\text{To}}_{\phi}$ and $ICC^{\text{Sh}}_{\phi}$.

\section{Mathematical properties of ICC}\label{Axioms}
This Section demonstrates that intrinsic causal contributions satisfy several desirable properties \citep{pmlr-v108-janzing20a}. 
To the best of our knowledge, our effort here is the first attempt at an axiomatic characterization of ICC.\footnote{\citet{do-shap} highlight the complete characterization of ICC as an open problem.} 
Due to space limitations, all proofs are provided in the Appendix \ref{Proof}.

\begin{theoremEnd}[end, restate, text link= , text proof= ]{proE}[Efficiency/ Completeness]\label{Efficiency}
\begin{align*}
 \sum_{j} ICC_{\phi}^{\text{To}}(X_{j}\rightarrow\hat{Y})= \phi(\hat{Y}|\mathbf{U})- \phi(\hat{Y}). 
\end{align*}
\end{theoremEnd}

In the context of neural network attribution, efficiency means that the uncertainty in the model’s output is fully distributed across its input features.
Shapley-based ICC values inherently guarantee the completeness, owing to the general properties of Shapley values 
\citep{Shapley+1953+307+318}.

\begin{proofE}
\begin{align*}
 \sum_{j} ICC_{\phi}^{\text{To}}(X_{j}\rightarrow\hat{Y}) & = \dfrac{1}{\lvert \mathcal{C}(\mathcal{G}) \rvert} \sum_{j} \sum_{\pi\in \mathcal{C}(\mathcal{G})} ICC_{\phi}^{\text{To}}(X_{j}\rightarrow\hat{Y}|I=T_{\pi}^{j})\\
 & = \dfrac{1}{\lvert \mathcal{C}(\mathcal{G}) \rvert} \sum_{\pi\in \mathcal{C}(\mathcal{G})} \sum_{j} ICC_{\phi}^{\text{To}}(X_{j}\rightarrow\hat{Y}|I=T_{\pi}^{j})\\
 & = \dfrac{1}{\lvert \mathcal{C}(\mathcal{G}) \rvert} \sum_{\pi\in \mathcal{C}(\mathcal{G})} \underbrace{\sum_{j} \phi(\hat{Y}|\mathbf{U}_{T_{\pi}^{j}\cup \{j\}})-\phi(\hat{Y}|\mathbf{U}_{T_{\pi}^{j}}) }_{\phi(\hat{Y}|\mathbf{U}_{V})- \phi(\hat{Y})}\\
 & = \dfrac{1}{\lvert \mathcal{C}(\mathcal{G}) \rvert} \cdot \lvert \mathcal{C}  (\mathcal{G}) \rvert \cdot \big(\phi(\hat{Y}|\mathbf{U}_{V})- \phi(\hat{Y})\big) = \phi(\hat{Y}|\mathbf{U})- \phi(\hat{Y})
\end{align*}
\end{proofE}

\begin{theoremEnd}[end, restate, text link=, text proof= ]{proE}[Nullity/ Dummy]\label{Nullity}
 \begin{align*}
 ICC_{\phi}^{\text{To}} (X_{j}\rightarrow\hat{Y}) = ICC_{\phi}^{\text{Sh}} (X_{j}\rightarrow\hat{Y}) = 0
 \end{align*}
 whenever $\phi(\hat{Y}|\mathbf{U}_{I})= \phi(\hat{Y}|\mathbf{U}_{I\cup \{ j\}})$ for all $I\subseteq [p]-\{j\}$.
\end{theoremEnd}

\begin{proofE}

 By definition, if $\phi(\hat{Y}|\mathbf{U}_{I})= \phi(\hat{Y}|\mathbf{U}_{I\cup \{ j\}})$, then $ICC_{\phi}(X_{j} \rightarrow \hat{Y}) = 0$. The result follows immediately when this holds for all $I \subseteq [p] - {j}$.
\end{proofE}

Nullity ensures that if a feature is entirely disconnected from the model’s output, it receives no contribution.

\begin{theoremEnd}[end, restate, text link=, text proof= ]{proE}[Symmetry]\label{Symmetry}
\begin{align*}
 ICC_{\phi}^{\text{Sh}} (X_{j}\rightarrow\hat{Y}) = ICC_{\phi}^{\text{Sh}} (X_{l}\rightarrow\hat{Y}), 
\end{align*}
 \text{if } $\phi(\hat{Y}|\mathbf{U}_{I\cup \{ j\}}) = \phi(\hat{Y}|\mathbf{U}_{I\cup \{ l\}})$ \: $\forall I$ \text{ s.t. } $I\subseteq [p]-\{j,l\}$. 
\end{theoremEnd}

\begin{proofE}

Note that for $I=\emptyset$, we have $\phi(\hat{Y}|U_{j}) = \phi(\hat{Y}|U_{l})$.
For any $I\subseteq [p]-\{j,l\}$, if $\phi(\hat{Y}|\mathbf{U}_{I\cup \{ j\}}) = \phi(\hat{Y}|\mathbf{U}_{I\cup \{ l\}})$, then it follows that $ICC_{\phi}(X_{j}\rightarrow\hat{Y}|I)=ICC_{\phi}(X_{l}\rightarrow\hat{Y}|I)$. 
More importantly, for each $T\subseteq [p]-\{j\}$ with $l\in T$, there exists a corresponding subset $T'\subseteq [p]-\{ l\}$ such that $j\in T'$ and
\begin{align*}
\phi(\hat{Y}|\mathbf{U}_{T})=\phi(\hat{Y}|\mathbf{U}_{T'}); \quad \phi(\hat{Y}|\mathbf{U}_{T\cup \{ j \}})=\phi(\hat{Y}|\mathbf{U}_{T' \cup \{ l\}}). 
\end{align*}
This can be seen from the following:
\begin{align*}
 \phi(\hat{Y}|\mathbf{U}_{T\cup \{ j \}})=\phi(\hat{Y}|\mathbf{U}_{T -\{ l\} },U_ {l},U_ {j} ) = \phi(\hat{Y}|\mathbf{U}_{\underbrace{T -\{ l\}\cup\{j\}}_{T'} },U_ {l} ) = \phi(\hat{Y}|\mathbf{U}_{T'\cup \{ l \}}),
\end{align*}
and since $T-\{l\} \subseteq [p]-\{l,j\}$, we have: 
\begin{align*}
\phi(\hat{Y}|\mathbf{U}_{T'}) = \phi(\hat{Y}|\mathbf{U}_{T-\{ l\}}, U_{j})=\phi(\hat{Y}|\mathbf{U}_{T-\{ l\}}, U_{l})= \phi(\hat{Y}|\mathbf{U}_{T}). 
\end{align*}
As a result, we get
\begin{align*}
 ICC_{\phi}(X_{j}\rightarrow\hat{Y}|T)= ICC_{\phi}(X_{l}\rightarrow\hat{Y}|T'),
\end{align*}
with $|T|=|T'|$. From the results above, we can easily deduce the following equality:
\begin{align*}
 ICC_{\phi}^{\text{Sh}}(X_{j}\rightarrow \hat{Y}) & = \sum_{T \subseteq [p]/\{ j\} } \dfrac{1}{n \binom{n-1}{|T|} } ICC_{\phi}(X_{j}\rightarrow \hat{Y}| T)\\ & = \sum_{T \subseteq [p]-\{ j,l\} } \dfrac{1}{n \binom{n-1}{|T|} } ICC_{\phi}(X_{j}\rightarrow \hat{Y}| T) +\sum_{\substack{T \subseteq [p]-\{j\} \\ l \in T }} \dfrac{1}{n \binom{n-1}{|T|} } ICC_{\phi}(X_{j}\rightarrow \hat{Y}| T) \\ & =
 \sum_{T \subseteq [p]-\{ j,l\} } \dfrac{1}{n \binom{n-1}{|T|} } ICC_{\phi}(X_{l}\rightarrow \hat{Y}| T) +\sum_{\substack{T' \subseteq [p]-\{l\} \\ j \in T' }} \dfrac{1}{n \binom{n-1}{|T'|} } ICC_{\phi}(X_{l}\rightarrow \hat{Y}| T') \\
 & = \sum_{T \subseteq [p]/\{ l\} } \dfrac{1}{n \binom{n-1}{|T|} } ICC_{\phi}(X_{l}\rightarrow \hat{Y}| T)\\ & = ICC_{\phi}^{\text{Sh}}(X_{l}\rightarrow \hat{Y})
\end{align*}
\end{proofE}
Symmetry requires that attribution be equally distributed among features that provide the same information for the model’s prediction. While $ICC_{\phi}^{\text{Sh}}$ satisfies the symmetry property, $ICC_{\phi}^{\text{To}}$ does not. However, the symmetry property is not without controversy, as symmetrical approaches to model explainability can obscure known causal relationships in the data \citep{frye2021shapley}.


\begin{theoremEnd}[end, restate, text link=, text proof= ]{proE}[Sensitivity/ Causal irrelevance]\label{Sensitivity}
If $X_{i}$ is causally irrelevant \citep{GALLES19979} to $\hat{Y}$ for all $I\subseteq [p]-\{i\}$, i.e., 
\begin{align*}
P(\hat{Y}|do(X_{i},\mathbf{X}_{I})) = P(\hat{Y}|do(\mathbf{X}_{I})), \quad \forall I \subseteq [p]-\{i\},
\end{align*}
 then 
\begin{align*}
 ICC^{\text{Sh}}_{\phi}(X_{i}\rightarrow\hat{Y}) = ICC^{\text{To}}_{\phi}(X_{i}\rightarrow\hat{Y}) = 0. 
\end{align*} 
\end{theoremEnd}

\begin{proofE}
Note that for $I=\emptyset$,
\begin{align}\label{sen_1}
 P(\hat{Y}|do(X_{i}=x_{i}))=P(\hat{Y}).
\end{align}
The possible (natural) values of $\hat{Y}$ are:
\begin{align*}
 \mathcal{Y} & = \{ f(x_{i}',\mathbf{x}_{-i})| x_{i}'\in \mathcal{X}_{i}, \mathbf{x}_{-i} \in \mathcal{X}_{-i} \}
 \\
 & = \{ f^{*}(u_{i},\mathbf{u}_{-i})| u_{i}\in \mathcal{U}_{i}, \mathbf{u}_{-i} \in \mathcal{U}_{-i} \},
\end{align*}
where $\mathcal{X}_{-i}$, $\mathcal{X}_{i}$, $\mathcal{U}_{-i}$ and $\mathcal{U}_{i}$ are the supports of $\mathbf{X}_{-i}, X_{i}, \mathbf{U}_{-i}$ and $U_{i}$, respectively. 
Similarly, under $do(X_{i}=\mathbf{x}_{i})$, the possible values of $\hat{Y}$ are given by:
\begin{align*}
 \mathcal{Y}_{x_{i}} & = \big\{ f(x_{i},\mathbf{x}_{-i})| \mathbf{x}_{-i}\in \mathcal{X}_{-i}^{do(x_{i})} \big\}\\
 & = \{ \Tilde{f}(x_{i},\mathbf{u}_{-i})|\mathbf{u}_{-i}\in \mathcal{U}_{-i}\},
\end{align*}
where $\mathcal{X}_{-i}^{do(x_{i})}$ is the support of $\mathbf{X}_{-i}$ under the intervention $do(X_{i}=x_{i})$. The equality of distributions in equation \ref{sen_1} 
 imposes the constraint that the support of 
$Y$ under the intervention must match its natural support, i.e., $\mathcal{Y}=\mathcal{Y}_{x_{i}}$. In other words, for each $(u_{i},\mathbf{u}_{-i})\in \mathcal{U}, \exists \mathbf{u}_{-i}' \in \mathcal{U}_{-i}$ such that $f^{*}(u_{i},\mathbf{u}_{-i}) =\Tilde{f}(x_{i}, \mathbf{u}_{-i}')$. Since $\mathbf{U}$ does not depend on $X_{i}$, the function $f^{*}$ also does not depend on $U_{i}$. Therefore, $\hat{Y}$ functionally does not depend on $U_{i}$. Thus, for any $I\subseteq [p]-\{ i\}$,
\begin{align*}
\phi(\hat{Y}|\mathbf{U}_{i+I})= \phi(f^{*}(\mathbf{U}_{-i})|\mathbf{U}_{I}, U_{i}) = \phi(f^{*}(\mathbf{U}_{-i})|\mathbf{U}_{I})= \phi(\hat{Y}|\mathbf{U}_{I}).
\end{align*}
The rest follows directly from Property \ref{Nullity}.

\end{proofE}
Causal irrelevance captures the causes of an outcome by ensuring that variables not related to the outcome have zero contribution.
From a causal viewpoint, it is also related to sensitivity \citep{IG}: if the function implemented by the deep network does not mathematically depend on a particular variable, then the attribution for that variable should always be zero. Implementation invariance \citep{IG} axiom loses significance if it refers to the properties of functions rather than focusing on the properties of algorithms \citep{pmlr-v108-janzing20a}. While linearity is often a desirable property in many attribution methods, recent progress has been towards non-linear attribution methods \citep{Look_at_variance}. The linearity of ICC depends on the choice of the function $\phi$. However, we sacrifice linearity by focusing on using a variance-based uncertainty measure as a candidate for $\phi$.

\paragraph{Choice of $\phi$.}
So far, we have considered $\phi$ as a general measure of uncertainty without specifying its form. However, we need to adopt a suitable $\phi$ for practical purposes.
While \cite{janzing2024quantifyingintrinsiccausalcontributions} suggest using variance and entropy for contribution analysis, we will focus on variance-based uncertainty measures in this article.
Quantifying uncertainty using variance is often more intuitive and easier to estimate from finite data. Furthermore, variance-based measures meet several desirable properties (axioms) for assessing second-order uncertainty \citep{corbiere:hal-03347628}, as discussed by \citet{sale2023second}. They have been proposed as an alternative to entropy-based measures, which have recently faced criticism in the literature\citep{pmlr-v216-wimmer23a}.
By defining $\phi(\hat{Y}|\mathbf{U}_{I}):= \mathbb{V}_{\mathbf{U}_{I}}(\mathbb{E}(\hat{Y}|\mathbf{U}_{I}))$, we can express the contribution of variable $X_j$ to $\hat{Y}$, given the context $I$, as:
\begin{align*}
 ICC_{\phi}(X_{j}\rightarrow \hat{Y}|I ) = \mathbb{V}_{\mathbf{U}_{I+j}}(\mathbb{E}(\hat{Y}|\mathbf{U}_{ I+j })) - \mathbb{V}_{\mathbf{U}_{I}}(\mathbb{E}(\hat{Y}|\mathbf{U}_{I})) 
\end{align*}
The difference between two variances allows us to measure the intrinsic contribution of $X_j$ to the uncertainty in predicting $\hat{Y}$, relative to the context $I$. 
The monotonicity of the variance of the conditional expectation is immediate from the following theorem.
\begin{theoremE}[][end, restate, text link=, text proof= ]\label{Thm1}
Let $X,Y$ and $Z$ be random variables on the same probability space and $\mathbb{V}(X)< \infty$. Then, 
 $$\mathbb{V}_{Y}(\mathbb{E}(X|Y)) \leq \mathbb{V}_{Y,Z}(\mathbb{E}(X|Y,Z)).$$
\end{theoremE}
\begin{proofE}
The law of total variance states that
$\mathbb{V}(X)= \mathbb{E}_{Y}(\mathbb{V}(X|Y)) + \mathbb{V}_{Y}(\mathbb{E}(X|Y)).$
Similarly, $\mathbb{V}(X)= \mathbb{E}_{Y,Z}(\mathbb{V}(X|Y,Z)) + \mathbb{V}_{Y,Z}(\mathbb{E}(X|Y,Z)).$ From this, it follows  that the expected variance of $X$ is greater than or equal to the expected value of the conditional variance of 
$X$ given $Y$, i.e., 
$\mathbb{E}(\mathbb{V}(X))\geq \mathbb{E}_{Y}(\mathbb{V}(X|Y))$ which also implies the conditional version $\mathbb{E}_{Z}(\mathbb{V}(X|Z))\geq \mathbb{E}_{Y,Z}(\mathbb{V}(X|Y,Z))$ for any random variable $Z$. Interchanging $Y$ and $Z$ in the last expression and subtracting the expected variances from $\mathbb{V}(X)$, we obtain the stated inequality. 
\end{proofE}
Note that while Theorem \ref{Thm1} is stated for random variables, it also holds for random vectors $\mathbf{Y},\mathbf{Z}$.
We have hitherto guaranteed that contributions are positive. However, normalized contributions are easier to interpret and visualize. The decomposition of the total variance of 
$\hat{Y}$ provides a natural way for normalizing the ICC:
\begin{corollary}[Causal Decomposition of Variance]\label{Causal_decompostion}
Let $\mathbb{V}(\hat{Y})< \infty$. Then, 
\begin{align*}
 \mathbb{V}(\hat{Y}) = \sum_{j=1}^{p} ICC_{\phi}^{\text{To}}(X_{j}\rightarrow \hat{Y}) = \dfrac{1}{\lvert \mathcal{C}(\mathcal{G}) \rvert} \sum_{j=1}^{p}\sum_{\pi\in \mathcal{C}(\mathcal{G})} \Big( \mathbb{V}\big(\mathbb{E}(\hat{Y}|\mathbf{X}_{T^{j}_{\pi}+j})\big) -\mathbb{V}\big(\mathbb{E}(\hat{Y}|\mathbf{X}_{T^{j}_{\pi}})\big) \Big) . 
\end{align*} 
\end{corollary}
The proof of the Corollary \ref{Causal_decompostion} is immediate from Property \ref{Efficiency} and our choice of $\phi$. A similar variance decomposition is straightforward for Shapley-based ICC. Going forward, we will consider 
$\phi$ to be normalized by $\mathbb{V}(\hat{Y})$. i.e., $\phi(\hat{Y}|\mathbf{U}_{I}):= \frac{\mathbb{V}_{\mathbf{U}_{I}}(\mathbb{E}(\hat{Y}|\mathbf{U}_{I}))}{\mathbb{V}(\hat{Y})}$.

\section{Comparison with Sobol’ method }\label{sobolsection}
While variance-based sensitivity analysis \citep{SOBOL} accommodates non-linear models, it falls short of capturing causal influence. This is because it focuses on reducing variance by conditioning on observed variables without distinguishing whether the statistical relationship with the target is causal or merely confounded.
We view corollary \ref{Causal_decompostion} as a causal decomposition of variance as it allows for the partial allocation of the output variance to each input variable while respecting the causal ordering, thereby generalizing classical functional ANOVA decomposition of variance within a causal framework. Although efforts \citep{LI, KUCHERENKO2012937, doi:10.1137/120904378, eb4f7804-e31a-3e9c-abf5-60bb5e0ffa06} have been made to generalize ANOVA by removing the independence assumption among input variables, none have addressed this issue from a causal perspective. Finally, we establish the connection between variance-based ICC and Sobol' indices, assuming independent input variables, in the following theorem:

\begin{theoremE}[][end, restate, text link=, text proof= ]\label{Sobol-ICC}
Assume that input features $\{X_{i}\}$ are independent. 
Then, with our specific choice of $\phi$, for any $I\subseteq[p]$ and any $j\in[p]$, we have 
\begin{align}
\phi(\hat{Y}|\mathbf{U}_{I})= \phi(\hat{Y}|\mathbf{X}_{I})=\sum_{T\subseteq I} \mathcal{S}_{T}; \qquad 
ICC_{\phi}(X_{j}\rightarrow \hat{Y}) =\sum_{j\in T, T\subseteq [p]} \dfrac{\mathcal{S}_{T}}{|T|},
\end{align}
where $\mathcal{S}_{T}$ is the Sobol' sensitivity index of the input subset $\mathbf{X}_{T}$ (see Appendix \ref{VSA} for more details).
\end{theoremE}

\begin{proofE}
When the input features are independent, we have $\mathcal{C(G)}=S_{p}$. Consequently, from Lemma \ref{Identifiability}, for any $\pi \in S_{p}$, the following holds: \begin{align*}
 \phi(\hat{Y}|\mathbf{U}_{T^{j}_{\pi}}) = \phi(\hat{Y}|\mathbf{X}_{T^{j}_{\pi}}). 
\end{align*}
In other words, for any subset $I\subseteq[p]$, 
$\phi(\hat{Y}|\mathbf{U}_{I}) = \phi(\hat{Y}|\mathbf{X}_{I}).$
For our specific choice of $\phi$, we have: 
\begin{align*}
 \phi(\hat{Y}|\mathbf{U}_{I}) = \phi(\hat{Y}|\mathbf{X}_{I})=\dfrac{\mathbb{V}_{\mathbf{X}_{I}}(\mathbb{E}(\hat{Y}|\mathbf{X}_{I}))}{\mathbb{V} (\hat{Y})} = \dfrac{\sum_{T\subseteq I} \mathbb{V}(f_{T}(\mathbf{X}_{T}))}{\mathbb{V}(f(\mathbf{X})} = \sum_{T\subseteq I}\mathcal{S}_{T}, 
\end{align*}
where the third equality follows directly from Equations \ref{anova2} and \ref{anova3}. 

 With independent input features,it follows from Equations \ref{ICC_shapley} and \ref{ICC_topological} that $ICC_{\phi}^{\text{To}}$ and $ICC_{\phi}^{\text{Sh}}$ coincide:
\begin{align*}
 ICC_{\phi}^{\text{To}} (X_{j}\rightarrow\hat{Y}) & = ICC_{\phi}^{\text{Sh}} (X_{j}\rightarrow\hat{Y})\\ & = \sum_{T \subseteq [p]/\{ j\} } \dfrac{1}{p \binom{p-1}{|T|} } ICC_{\phi}(X_{j}\rightarrow \hat{Y}| T) \\ 
 & = \sum_{T \subseteq [p]/\{ j\} } \dfrac{1}{p \binom{p-1}{|T|} } \Big(\phi(\hat{Y}|\mathbf{U}_{j + T}) - \phi(\hat{Y}|\mathbf{U}_{T})\Big)\\
 & = \sum_{T \subseteq [p]/\{ j\} } \dfrac{1}{p \binom{p-1}{|T|} } \Big(\phi(\hat{Y}|\mathbf{X}_{j + T}) - \phi(\hat{Y}|\mathbf{X}_{T})\Big)
 \\ & = \sum_{T \subseteq [p]/\{ j\} } \dfrac{1}{p \binom{p-1}{|T|} } \Big( \mathbb{V}(\mathbb{E}(f(\mathbf{X})|\mathbf{X}_{T+j})) - \mathbb{V}(\mathbb{E}(f(\mathbf{X})|\mathbf{X}_{T})) \Big), 
\end{align*} The rest of the proof is immediate from Theorem \ref{Sobol_shapley_thm}. 
\end{proofE}

Theorem \ref{Sobol-ICC} reflects that ICC may be viewed as a step toward generalizing Sobol indices within a causal framework.

\section{Learning and explaining
ICC in NNs}

In this Section, we introduce our primary post-hoc \citep{RETZLAFF2024101243} approach to identifying and explaining the intrinsic causal contributions of an input feature. We will start by formally defining the concept of identifiability within the context of ICC.
\begin{definition}\label{identifiability_definition}
For a given neural network $\mathcal{N}$,
the intrinsic causal contribution of an input feature $X_{j}$ on the output $\hat{Y}$ is identifiable if $ICC_{\phi}(X_{j}\rightarrow \hat{Y})$ can be computed \textit{uniquely} from any positive probability distribution $P(\mathbf{X},\hat{Y})$. 
\end{definition}
Prior work \citep{janzing2024quantifyingintrinsiccausalcontributions}  does not address the issue of identifiability for ICC. 
Under the assumption of no latent confounding (Assumption \ref{assumption1}), and based on Lemma \ref{Identifiability}, it is straightforward that $ICC_{\phi}^{\text{To}}$ is identifiable. However, in general, we are unable to find a way to compute $\phi(\hat{Y}|\mathbf{U}_{I})$ without knowledge of the SCM. For example, \cite{janzing2024quantifyingintrinsiccausalcontributions} inferred the SCM based on common sense knowledge and assigned all regression coefficients a value of 1. For a dataset \citep{auto_mpg_9} with non-linearities, they applied an additive noise model for a simple approximation of structural equations. In contrast, we propose to learn the entire causal-generating process using causal normalizing flow (CNF) \citep{javaloy2023causal} as they are a natural choice for approximating a wide range of causal data-generating processes. Nevertheless, generative models are vulnerable to cases where the latent values ($\mathbf{u}$) underlying observations cannot be determined uniquely \citep{Khemakhem2019VariationalAA}, no matter how much empirical data is available, which may lead to inaccurate estimation of $\phi(\hat{Y}|\mathbf{U}_{I})$. In this scenario, we guarantee that, even though different but equivalent model (CNF) fits may be obtained from the same data, the estimation of $\phi$ remains consistent, provided the following assumptions hold. For further details on normalizing flows, see \cite{NF}.


\begin{assumption}\label{assumption_2}
We constrain the class of SCMs under consideration by adopting the following fairly common assumptions from \cite{javaloy2023causal}: i) the data-generating process is diffeomorphic — that is, $\mathbf{F}$ is invertible, and both 
$\mathbf{F}$ and its inverse are differentiable; ii) causal
sufficiency, i.e., $P_{\mathbf{U}}(\mathbf{u})=\prod_{j=1}^{p}P_{U_{j}}(u_{j})$.
\end{assumption}

Causal normalizing flows are themselves parametric TMI maps that can approximate any other TMI map with arbitrary precision. With SCMs and causal NFs categorized under the same family, we leverage existing results on identifiability \citep{DBLP:conf/aistats/XiB23}.

\begin{theorem}[\cite{javaloy2023causal}, \cite{DBLP:conf/aistats/XiB23}]\label{identify}
If two elements from the family $\mathcal{F}\times \mathcal{P_{U}} $ yield the same observational distribution, then their data-generating processes differ only by a component-wise (Borel measurable) invertible transformation of the variables $\mathbf{U}$.
\end{theorem}

Theorem \ref{identify} says that if a causal normalizing flow $(\mathbf{F}_{\theta},P_{\theta}) \in \mathcal{F}\times \mathcal{P_{U}} $ matches the observational distribution generated by $(\mathbf{F},P_{\mathbf{U}})\in\mathcal{F}\times \mathcal{P_{U}}$, 
then the exogenous variables in the flow differ from the true exogenous variables only through independent, component-wise invertible transformations. Mathematically, for $\mathbf{U}\sim P_{\mathbf{U}}$, it holds that $\mathbf{F}_{\theta}^{-1}(\mathbf{F}(\mathbf{U})) \sim P_{\theta} $ and 
$\mathbf{F}_{\theta}^{-1}(\mathbf{F}(\mathbf{u}))= \mathbf{h}(\mathbf{u}) = (h_{1}(u_{1}), h_{2}(u_{2}),...,h_{p}(u_{p}))$, where each $h_{i}$ is an invertible function. This component-wise invertibility is fundamental to the identifiability of ICC:

\begin{theoremE}[][end, restate, text link=, text proof= ]\label{ICC-identifiabilty}
Under  Assumption \ref{assumption_2}, suppose we have two CNFs $(\mathbf{F}_{\theta_{1}},P_{\theta_{1}})$ and $(\mathbf{F}_{\theta_{2}},P_{\theta_{2}})$ that both match the observational distribution $P(\mathbf{X},\hat{Y})$, then the
 intrinsic causal contributions of $X_{j}$ on $\hat{Y}$ will be equal for both CNFs. Specifically, we have:
 \begin{align*}
 ICC_{\phi, \theta_{1}}(X_{j}\rightarrow\hat{Y}) = ICC_{\phi, \theta_{2}}(X_{j}\rightarrow\hat{Y}). 
 \end{align*}
\end{theoremE}

\begin{proofE}
Continuing from Theorem \ref{identify}, 
since $h$ is measurable bijection, the $\sigma-$algebras generated by $u_{p}$ and $h_{p}(u_{p})$ are identical and thus we have $\mathbb{E}(\hat{Y}|u_{p}) = \mathbb{E}(\hat{Y}|h_{p}(u_{p}))$ \citep{MTPT}. More generally, for any subset $I\subseteq[p]$, it follows that 
$\mathbb{E}(\hat{Y}|\mathbf{u}_{I}) = \mathbb{E}(\hat{Y}|h_{I}(\mathbf{u}_{I}))$, where $h(\mathbf{u}_{I})=h_{I}(\mathbf{u}_{I})= \big(h_{j}(u_{j})\big)_{j\in I}$. 
Given our specific choice of $\phi$, it follows directly that $\phi(\hat{Y}|h(\mathbf{U}_{I}))= \phi(\hat{Y}|\mathbf{U}_{I})$, thereby establishing the equality
\begin{align}\label{independence}
 ICC_{\phi,\theta}(X_{j}\rightarrow\hat{Y}) =ICC_{\phi}(X_{j}\rightarrow\hat{Y}). 
\end{align}
The dependence on $\theta$ on the right-hand side arises from the expression
$
\mathbf{F}_{\theta}^{-1}(\mathbf{F}(\mathbf{u}))= \mathbf{h}(\mathbf{u}). 
$
Equation \ref{independence} states that if a CNF matches the observational distribution, then the ICC computed with respect to the flow does not depend on the flow parameter $\theta$.
The statement of the theorem follows immediately as a direct consequence.
\end{proofE}

Another crucial advantage of using CNF framework \citep{javaloy2023causal} is its ability to handle both mixed continuous-discrete data and partial knowledge of the causal graph, making it highly applicable to real-world scenarios. To handle discrete data, we adopt the general method by \cite{DBLP:conf/aistats/XiB23} that transforms the observed discrete variables into continuous ones by adding independent noise $\epsilon\in [0,1]$ —such as standard uniform noise — ensuring the original distribution remains recoverable. Essentially, we posit that discrete variables represent the integer parts of noisy continuous variables generated under an SCM that meets our assumptions. Thereby, it allows our theoretical and practical insights to remain applicable. Recently, \citet{SAFE}  have raised questions about using likelihood loss to train normalizing flows. In line with their approach, we use MMD loss instead of likelihood in our experiment to train the normalizing flows. In training the flow model, we focus on the crucial step of estimating $\phi$, which is essential for computing the intrinsic causal contributions.The computation of $\mathbb{V}_{\mathbf{U}_{I}}(\mathbb{E}_{\mathbf{U}_{-I}}(\hat{Y}|\mathbf{U}_{I}))$ involves a two-fold integration, which could be challenging. We therefore present a Monte Carlo-based algorithm for the efficient estimation of $\phi$.




\begin{algorithm}[H]
\caption{Pseudocode for estimating $\phi$}
\label{estimation_phi}
\begin{algorithmic}[1] 
 \Statex \textbf{Input:} Batch size $B$, context $I$ for conditioning, trained CNF $(\mathbf{F}_{\theta},P_{\theta})$, the neural network $\mathcal{N}$

 \State $\mathbf{u}_{M}^{(i)}, \mathbf{u}_{N}^{(i)}\sim P_{\theta} $ for $i=1,2,...,B.$ 
 \State $\mathbf{u}_{Q}^{(i)} = \Big( \mathbf{u}_{M_{-I}}^{(i)},\mathbf{u}_{N_{I}}^{(i)} \Big) $ for $i=1,2,...,B.$

 \State $\hat{y}_{M}^{(i)} = \mathcal{N}\Big(\underbrace{\mathbf{F}_{\theta}(\mathbf{u}_{M}^{(i)})}_{\mathbf{x}_{M}}\Big), \hat{y}_{N}^{(i)} = \mathcal{N}\Big(\underbrace{\mathbf{F}_{\theta}(\epsilon_{V})}_{\mathbf{x}_{N}}\Big), \hat{y}_{Q}^{(i)} = \mathcal{N}\Big(\underbrace{\mathbf{F}_{\theta}(\epsilon_{W})}_{\mathbf{x}_{Q}}\Big)$ for $i=1,2,...,B.$ 

 \State $\Bar{y} = \dfrac{1}{2B}\sum_{i=1}^{B} \Big(\hat{y}_{M}^{(i)}+\hat{y}_{N}^{(i)}\Big)$; $V = \dfrac{1}{2B-1}\sum_{i=1}^{B}\Big((\hat{y}_{M}^{(i)}-\Bar{y})^{2}+(\hat{y}_{N}^{(i)}-\Bar{y})^{2}\Big)$
 
 \State $\hat{\psi} = V - \dfrac{1}{2B}\sum_{i=1}^{B} \big(\hat{y}_{N}^{(i)} - \hat{y}_{Q}^{(i)}\big)^{2}$
 
 \Statex \textbf{Output:} $\hat{\phi}=\dfrac{\hat{\psi}}{V}$
\end{algorithmic}
\end{algorithm}

In Algorithm \ref{estimation_phi}, we employ the Jansen estimator \citep{JANSEN199935}, widely recognized as one of the most efficient \citep{Puy_2022}. Jansen’s method is commonly employed alongside a Monte Carlo sampling strategy. We improve upon the standard Monte Carlo method by employing a Randomized Quasi-Monte Carlo (RQMC) sampling strategy \citep{RQMC_pierre}, which generates low-discrepancy sample sequences for faster and more stable convergence rates \citep{L’Ecuyer2002,GERBER2015798}. RQMC methods enable them to be considered variance reduction techniques for the standard Monte Carlo method. Scrambled nets, a type of RQMC, offer valuable robustness properties \citep{LLN_scrambled}.
Our experiments utilize the most commonly used QMC method: Sobol' sequences \citep{SOBOL196786}, which can be scrambled \citep{Scrambling_Sobol}. Although Algorithm \ref{estimation_phi} could be applied to both $ICC^{\text{To}}$ and 
$ICC^{\text{Sh}}$, for the sake of completeness, we provide an algorithm in the Appendix \ref{algo_appendix} — using Lemma \ref{Identifiability} — specifically dedicated to computing $\phi$ for $ICC^{\text{To}}$, where the CNF is not necessary. 

\section{Experiment and analysis}
\begin{figure}[!ht] 
 \centering 
 \includegraphics[width=13cm, height=4cm]{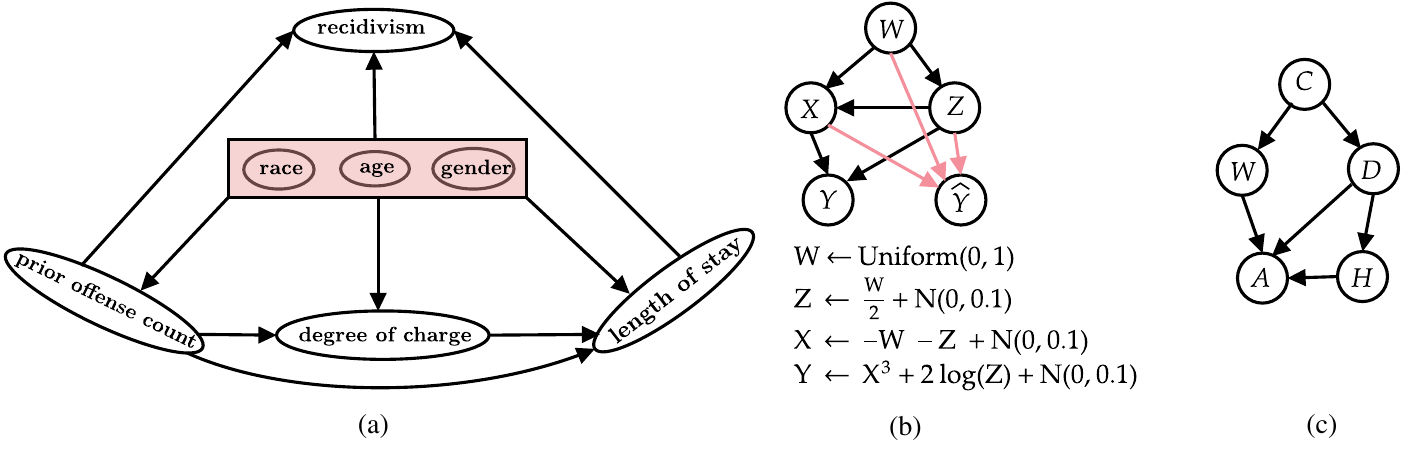}
 \caption{Causal graphs for experimental datasets: (a) COMPAS, (b) Synthetic, (c) AutoMPG }
 \label{causal_graphs_experiments} 
\end{figure}

Now, we demonstrate that ICCs provide a natural framework for global explanations.
We perform experiments on three datasets: a synthetic dataset and two well-known real-world benchmark datasets, AutoMPG \citep{auto_mpg_9} and COMPAS \citep{compas_data}. The causal graph of these datasets is depicted in Figure \ref{causal_graphs_experiments}. Appendix \ref{appendix_results} shows more detailed information on each dataset. We compare the ICC with global attributions generated by GAM, SP-LIME, and permutation feature importance (PFI). We apply GAM to five different local attribution methods: Integrated Gradients (IG), Gradient $\times$ Inputs (I$\times$G), SmoothGrad (SG), Shapley Values (SHAP), and LIME --- to generate global attributions for the test samples. We train a three-layer feed-forward neural network with ReLU activation functions on each of the three balanced datasets. The performance metrics of these networks are presented in Table \ref{Performance_table}. To calculate the ICC for each dataset, we fit a CNF to approximate the SCM of the input features. Each CNF is constructed using Masked Autoregressive Flows \citep{NIPS2017_6c1da886} as its layers.
We assess the quality of these flows using the 1-Wasserstein distance metric, with the results reported in Table \ref{Performance_table}.
We compute attribution scores on the test dataset. 
We use the OpenXAI \citep{agarwal2022openxai} codebase as the foundation for our implementation.



\begin{table}[h!]
\centering
\caption{\textbf{Left:} Performance metrics for $\mathcal{N}$ to Be Explained (Root Mean Squared Error (RMSE) for Regression and $F_{1}$ score for Classification). \textbf{Right:} Quality metrics (1-Wasserstein Distance) of CNF models used to compute the ICC.}
\label{tab:comparison_models}
\begin{minipage}{0.47\textwidth}
\centering
\begin{tabular}{lcc}
\toprule
$\mathcal{N}$ & \textbf{RMSE}($\downarrow$) & $\mathbf{F_{1}}$\textbf{ Score} ($\uparrow$) \\
\midrule
Synthetic & $0.1024_{\pm 0.0019}$ & N/A \\
Auto-MPG & $0.1103_{\pm 0.0032}$ & N/A \\
COMPAS & N/A & $0.9115_{\pm 0.0016}$ \\
\bottomrule
\end{tabular}
\end{minipage}%
\hspace{0.05\textwidth} 
\begin{minipage}{0.47\textwidth}
\centering
\begin{tabular}{lcc}
\toprule
\textbf{CNF Model} & $\mathcal{W}_{1}$-\textbf{Distance}\\
\midrule
Synthetic & $0.5553_{\pm0.0028}$\\
Auto-MPG &$ 0.9641_{\pm 0.0114}$\\
COMPAS & $0.9562_{\pm 0.0398}$\\
\bottomrule
\end{tabular}
\end{minipage}
\label{Performance_table}
\end{table}

Global explanation research inherently struggles with effective and reliable validation due to the absence of baseline truths for attributions, making identifying appropriate validation methodologies an open research question. 
In the absence of a ground truth, we evaluate the reliability of an attribution method by adapting the Prediction Gap on Unimportant feature perturbation (PGU) \citep{10.1145/3514094.3534159,DBLP:conf/bmvc/PetsiukDS18}. This metric measures the change in the network's output when unimportant features are set to zero, while the (top-$k$) influential features identified by a post hoc explanation remain unchanged. Smaller values on this metric indicate higher reliability in the explanation. For each dataset, the PGU values for every attribution method are reported in Table \ref{metric_table}.


\begin{figure}[h!]
 \centering
 \begin{minipage}{0.5\textwidth}
 \centering
 \includegraphics[width=\linewidth]{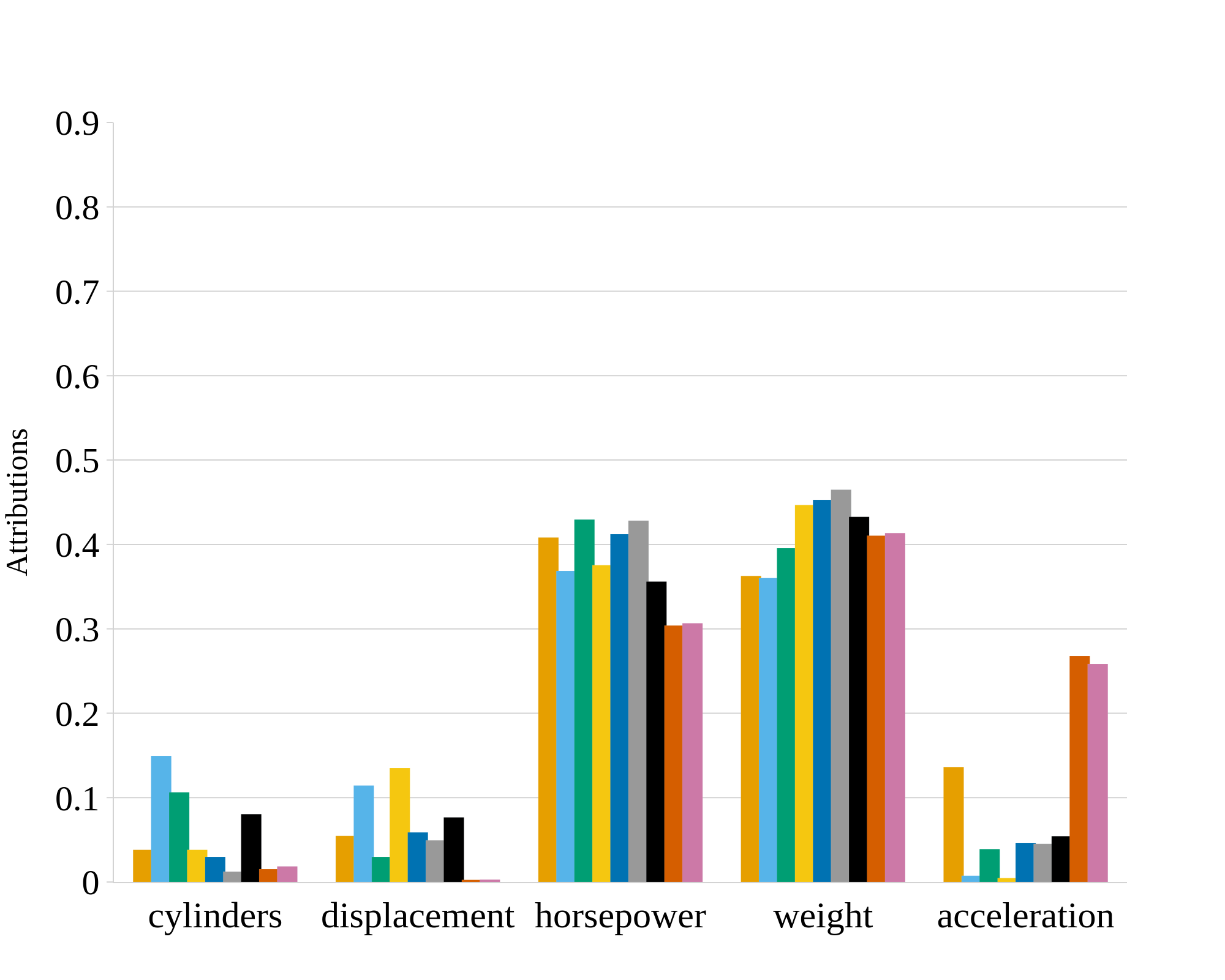}
 \end{minipage}\hfill
 \begin{minipage}{0.5\textwidth}
 \centering
\includegraphics[width=1\linewidth]{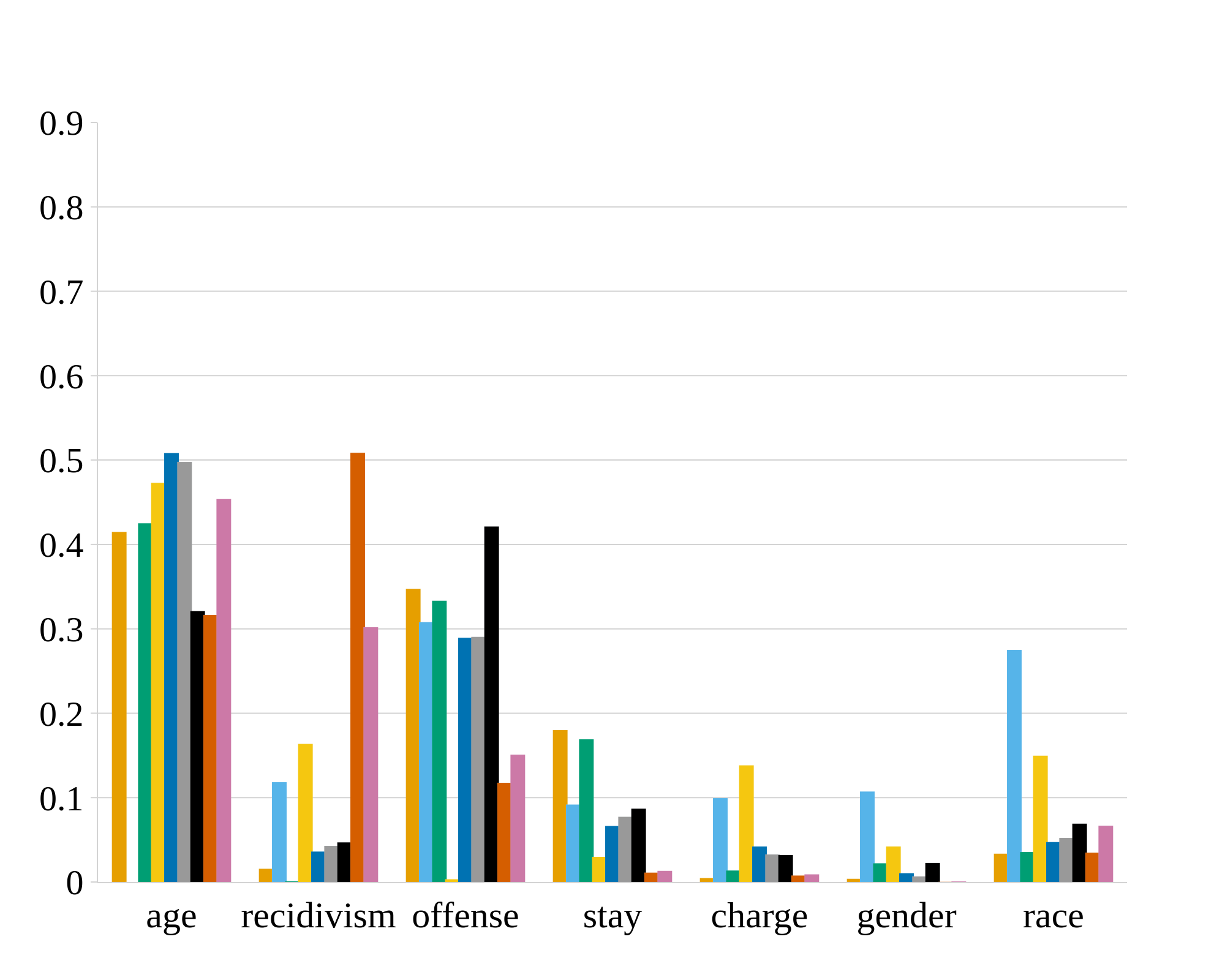}
 \end{minipage}\hfill
 \begin{minipage}{0.5\textwidth}
 \centering
\includegraphics[width=1\linewidth]{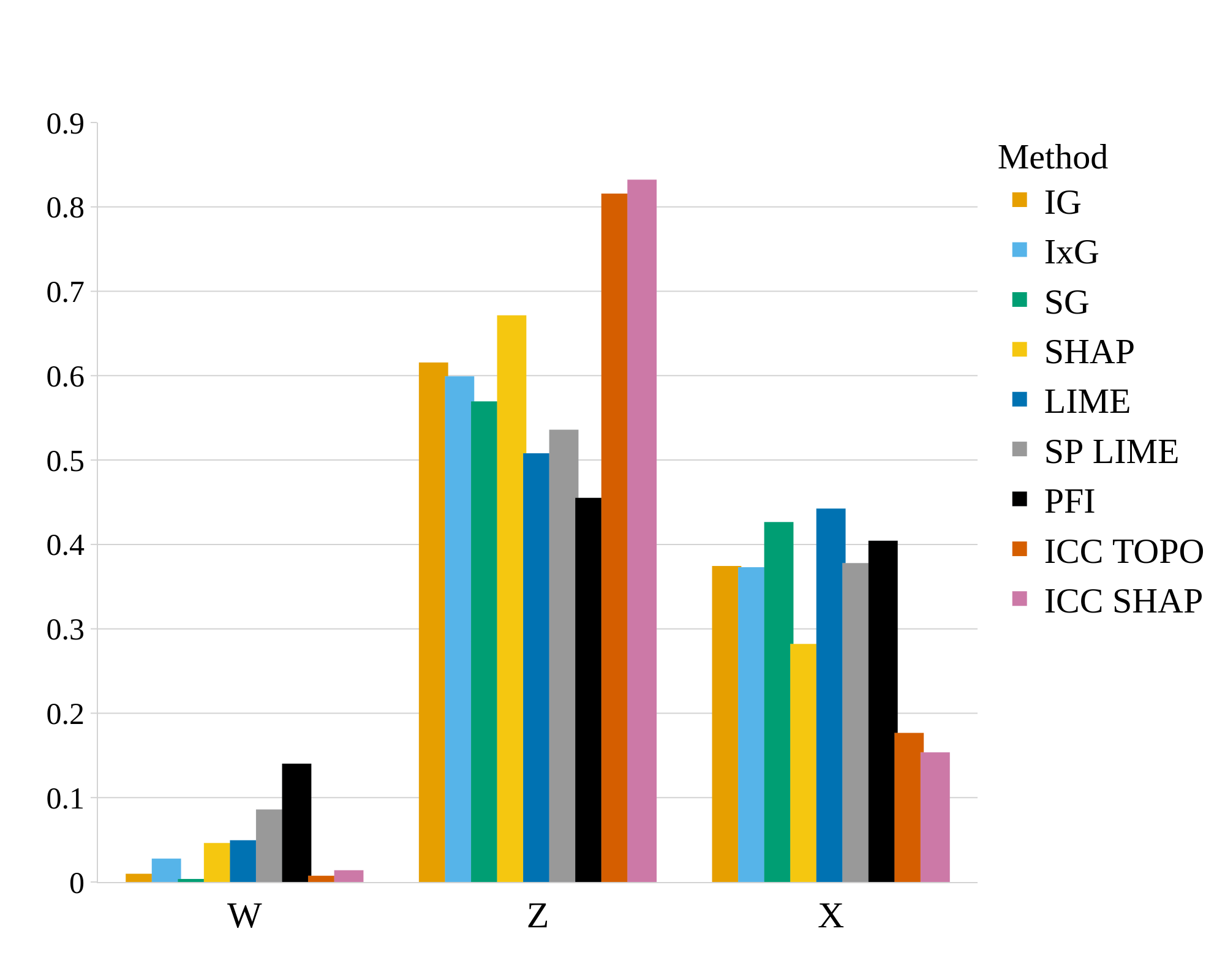}
 \end{minipage}
 \caption{Global attribution explanations - feature importances. Top left: AutoMPG dataset. Top right: COMPAS dataset. Bottom: Synthetic dataset.} 
 \label{attribution_values}
\end{figure}

Figure \ref{attribution_values} presents the attribution values of the input features. In the synthetic dataset, we observe that both ICC methods assign negligible attribution to 
$W$, which aligns with expectations as the true outcome 
$Y$ depends on $W$ only through 
$X$ and $Z$. SmoothGrad and IG methods also support this assignment. Similarly, in the AutoMPG experiment, both ICC methods produce comparable attributions.
However, a discrepancy appears in the COMPAS experiment between $ICC^{\text{To}}$ and $ICC^{\text{Sh}}$. Specifically, $ICC^{\text{To}}$ identifies recidivism as the most critical feature, while $ICC^{\text{Sh}}$ ranks it as the second most important. More notably, most other methods assign relatively low importance (attribution 
$\leq 0.25$) to recidivism, focusing instead on prior offense count as the more influential attribute. To further examine this discrepancy, we train separate classifiers using each feature individually: age, recidivism, and prior offense count. The resulting 
$F_{1}$ scores are 0.8972, 0.8964, and 0.8912, respectively.




\begin{table}[h!]
\centering
\caption{PGU($\downarrow$) values for different datasets (scaled by $1 \times 10^{-1}$). We report the aggregated PGU by summing across all values of 
$k$.
}
\label{tab:pgi_methods_transposed}
\begin{tabular}{|c|c|c|c|c|c|c|c|c|c|c|}
\hline
\rowcolor{lightgray} 
\textbf{Dataset} & \textbf{IG} & \textbf{I$\times$G} & \textbf{SG} & \textbf{SHAP} & \textbf{LIME} & \textbf{SP-LIME} & \textbf{PFI} & $\mathbf{ICC^{\text{To}}}$ & $\mathbf{ICC^{\text{Sh}}}$ \\
\hline
\hline
\cellcolor{lightgray}\textbf{Synthetic} & 1.9 & 1.9 & 1.9 & 1.9 & 1.9 & 4.77 & 1.9 & 1.9 & 1.9 \\

\hline
\cellcolor{lightgray}\textbf{Auto MPG} & 2.84 & 3 & 2.86 & 2.78 & 2.72 & 6.48 & 2.76 & \textbf{2.52} & \textbf{2.52} \\

\hline
\cellcolor{lightgray}\textbf{COMPAS} & 5.73 & 18.29 & 6.47 & 8.51 & 6.06 & 6.36 & 5.48 & 5.18 & \textbf{5.17} \\

\hline
\end{tabular}
\label{metric_table}
\end{table}

\section{Conclusion}
This paper proposes a framework that leverages Intrinsic Causal Contributions to generate global attributions that complement existing interpretability techniques for neural networks. Additionally, we establish a link between classical sensitivity analysis and Intrinsic Causal Contributions that bridge causality and sensitivity analysis.
This connection suggests a promising overlap area that warrants further research exploration.

\acks{This research was funded in part
by the Indo-French Centre for the Promotion of
Advanced Research (IFCPAR/CEFIPRA) through
project number CSRP 6702-2.}

\bibliography{refs}

\begin{thebibliography}{91}
\providecommand{\natexlab}[1]{#1}
\providecommand{\url}[1]{\texttt{#1}}
\expandafter\ifx\csname urlstyle\endcsname\relax
  \providecommand{\doi}[1]{doi: #1}\else
  \providecommand{\doi}{doi: \begingroup \urlstyle{rm}\Url}\fi

\bibitem[Agarwal et~al.(2022)Agarwal, Krishna, Saxena, Pawelczyk, Johnson, Puri, Zitnik, and Lakkaraju]{agarwal2022openxai}
Chirag Agarwal, Satyapriya Krishna, Eshika Saxena, Martin Pawelczyk, Nari Johnson, Isha Puri, Marinka Zitnik, and Himabindu Lakkaraju.
\newblock Open{XAI}: Towards a transparent evaluation of model explanations.
\newblock In \emph{Thirty-sixth Conference on Neural Information Processing Systems Datasets and Benchmarks Track}, 2022.
\newblock URL \url{https://openreview.net/forum?id=MU2495w47rz}.

\bibitem[Alvarez-Melis and Jaakkola(2017)]{alvarez-melis-jaakkola-2017-causal}
David Alvarez-Melis and Tommi Jaakkola.
\newblock A causal framework for explaining the predictions of black-box sequence-to-sequence models.
\newblock In Martha Palmer, Rebecca Hwa, and Sebastian Riedel, editors, \emph{Proceedings of the 2017 Conference on Empirical Methods in Natural Language Processing}, pages 412--421, Copenhagen, Denmark, September 2017. Association for Computational Linguistics.
\newblock \doi{10.18653/v1/D17-1042}.
\newblock URL \url{https://aclanthology.org/D17-1042}.

\bibitem[Athreya and Lahiri(2006)]{MTPT}
Krishna~B. Athreya and Soumen~N. Lahiri.
\newblock \emph{Measure Theory and Probability Theory (Springer Texts in Statistics)}.
\newblock Springer-Verlag, Berlin, Heidelberg, 2006.
\newblock ISBN 038732903X.

\bibitem[B{\'e}nesse et~al.(2024)B{\'e}nesse, Gamboa, Loubes, and Boissin]{Benesse2024-sx}
Cl{\'e}ment B{\'e}nesse, Fabrice Gamboa, Jean-Michel Loubes, and Thibaut Boissin.
\newblock Fairness seen as global sensitivity analysis.
\newblock \emph{Machine Learning}, 113\penalty0 (5):\penalty0 3205--3232, May 2024.

\bibitem[Bl{\"o}baum et~al.(2024)Bl{\"o}baum, G{\"o}tz, Budhathoki, Mastakouri, and Janzing]{blobaum2024dowhy}
Patrick Bl{\"o}baum, Peter G{\"o}tz, Kailash Budhathoki, Atalanti~A Mastakouri, and Dominik Janzing.
\newblock Dowhy-gcm: An extension of dowhy for causal inference in graphical causal models.
\newblock \emph{Journal of Machine Learning Research}, 25\penalty0 (147):\penalty0 1--7, 2024.

\bibitem[Breiman(2001)]{Breiman2001-mn}
Leo Breiman.
\newblock Random forests.
\newblock \emph{Machine Learning}, 45\penalty0 (1):\penalty0 5--32, October 2001.

\bibitem[Breuer et~al.(2024)Breuer, Sauter, Mohammadi, and Acar]{CAGE}
Nils~Ole Breuer, Andreas Sauter, Majid Mohammadi, and Erman Acar.
\newblock Cage: Causality-aware shapley value for global explanations.
\newblock In Luca Longo, Sebastian Lapuschkin, and Christin Seifert, editors, \emph{Explainable Artificial Intelligence}, pages 143--162, Cham, 2024. Springer Nature Switzerland.
\newblock ISBN 978-3-031-63800-8.

\bibitem[Brown et~al.(2020)Brown, Mann, Ryder, Subbiah, Kaplan, Dhariwal, Neelakantan, Shyam, Sastry, Askell, Agarwal, Herbert-Voss, Krueger, Henighan, Child, Ramesh, Ziegler, Wu, Winter, Hesse, Chen, Sigler, Litwin, Gray, Chess, Clark, Berner, McCandlish, Radford, Sutskever, and Amodei]{NEURIPS2020_1457c0d6}
Tom Brown, Benjamin Mann, Nick Ryder, Melanie Subbiah, Jared~D Kaplan, Prafulla Dhariwal, Arvind Neelakantan, Pranav Shyam, Girish Sastry, Amanda Askell, Sandhini Agarwal, Ariel Herbert-Voss, Gretchen Krueger, Tom Henighan, Rewon Child, Aditya Ramesh, Daniel Ziegler, Jeffrey Wu, Clemens Winter, Chris Hesse, Mark Chen, Eric Sigler, Mateusz Litwin, Scott Gray, Benjamin Chess, Jack Clark, Christopher Berner, Sam McCandlish, Alec Radford, Ilya Sutskever, and Dario Amodei.
\newblock Language models are few-shot learners.
\newblock In H.~Larochelle, M.~Ranzato, R.~Hadsell, M.F. Balcan, and H.~Lin, editors, \emph{Advances in Neural Information Processing Systems}, volume~33, pages 1877--1901. Curran Associates, Inc., 2020.
\newblock URL \url{https://proceedings.neurips.cc/paper_files/paper/2020/file/1457c0d6bfcb4967418bfb8ac142f64a-Paper.pdf}.

\bibitem[Chattopadhyay et~al.(2019)Chattopadhyay, Manupriya, Sarkar, and Balasubramanian]{pmlr-v97-chattopadhyay19a}
Aditya Chattopadhyay, Piyushi Manupriya, Anirban Sarkar, and Vineeth~N Balasubramanian.
\newblock Neural network attributions: A causal perspective.
\newblock In Kamalika Chaudhuri and Ruslan Salakhutdinov, editors, \emph{Proceedings of the 36th International Conference on Machine Learning}, volume~97 of \emph{Proceedings of Machine Learning Research}, pages 981--990. PMLR, 09--15 Jun 2019.
\newblock URL \url{https://proceedings.mlr.press/v97/chattopadhyay19a.html}.

\bibitem[Corbi{\`e}re et~al.(2021)Corbi{\`e}re, Lafon, Thome, Cord, and P{\'e}rez]{corbiere:hal-03347628}
Charles Corbi{\`e}re, Marc Lafon, Nicolas Thome, Matthieu Cord, and Patrick P{\'e}rez.
\newblock {Beyond First-Order Uncertainty Estimation with Evidential Models for Open-World Recognition}.
\newblock In \emph{{ICML 2021 Workshop on Uncertainty and Robustness in Deep Learning}}, Virtual, Austria, September 2021.
\newblock URL \url{https://cnam.hal.science/hal-03347628}.

\bibitem[Dai et~al.(2022)Dai, Upadhyay, Aivodji, Bach, and Lakkaraju]{10.1145/3514094.3534159}
Jessica Dai, Sohini Upadhyay, Ulrich Aivodji, Stephen~H. Bach, and Himabindu Lakkaraju.
\newblock Fairness via explanation quality: Evaluating disparities in the quality of post hoc explanations.
\newblock In \emph{Proceedings of the 2022 AAAI/ACM Conference on AI, Ethics, and Society}, AIES '22, page 203–214, New York, NY, USA, 2022. Association for Computing Machinery.
\newblock ISBN 9781450392471.
\newblock \doi{10.1145/3514094.3534159}.
\newblock URL \url{https://doi.org/10.1145/3514094.3534159}.

\bibitem[Dandl et~al.(2020)Dandl, Molnar, Binder, and Bischl]{10.1007/978-3-030-58112-1_31}
Susanne Dandl, Christoph Molnar, Martin Binder, and Bernd Bischl.
\newblock Multi-objective counterfactual explanations.
\newblock In Thomas B{\"a}ck, Mike Preuss, Andr{\'e} Deutz, Hao Wang, Carola Doerr, Michael Emmerich, and Heike Trautmann, editors, \emph{Parallel Problem Solving from Nature -- PPSN XVI}, pages 448--469, Cham, 2020. Springer International Publishing.
\newblock ISBN 978-3-030-58112-1.

\bibitem[Dash et~al.(2022)Dash, Balasubramanian, and Sharma]{dash2022evaluating}
Saloni Dash, Vineeth~N Balasubramanian, and Amit Sharma.
\newblock Evaluating and mitigating bias in image classifiers: A causal perspective using counterfactuals.
\newblock In \emph{Proceedings of the IEEE/CVF Winter Conference on Applications of Computer Vision}, pages 915--924, 2022.

\bibitem[FEL et~al.(2021)FEL, Cadene, Chalvidal, Cord, Vigouroux, and Serre]{Look_at_variance}
Thomas FEL, Remi Cadene, Mathieu Chalvidal, Matthieu Cord, David Vigouroux, and Thomas Serre.
\newblock Look at the variance! efficient black-box explanations with sobol-based sensitivity analysis.
\newblock In M.~Ranzato, A.~Beygelzimer, Y.~Dauphin, P.S. Liang, and J.~Wortman Vaughan, editors, \emph{Advances in Neural Information Processing Systems}, volume~34, pages 26005--26014. Curran Associates, Inc., 2021.
\newblock URL \url{https://proceedings.neurips.cc/paper_files/paper/2021/file/da94cbeff56cfda50785df477941308b-Paper.pdf}.

\bibitem[Frosst and Hinton(2017)]{Frosst2017DistillingAN}
Nicholas Frosst and Geoffrey~E. Hinton.
\newblock Distilling a neural network into a soft decision tree.
\newblock \emph{ArXiv}, abs/1711.09784, 2017.
\newblock URL \url{https://api.semanticscholar.org/CorpusID:3976789}.

\bibitem[Frye et~al.(2021)Frye, de~Mijolla, Begley, Cowton, Stanley, and Feige]{frye2021shapley}
Christopher Frye, Damien de~Mijolla, Tom Begley, Laurence Cowton, Megan Stanley, and Ilya Feige.
\newblock Shapley explainability on the data manifold.
\newblock In \emph{International Conference on Learning Representations}, 2021.
\newblock URL \url{https://openreview.net/forum?id=OPyWRrcjVQw}.

\bibitem[Galles and Pearl(1997)]{GALLES19979}
David Galles and Judea Pearl.
\newblock Axioms of causal relevance.
\newblock \emph{Artificial Intelligence}, 97\penalty0 (1):\penalty0 9--43, 1997.
\newblock ISSN 0004-3702.
\newblock \doi{https://doi.org/10.1016/S0004-3702(97)00047-7}.
\newblock URL \url{https://www.sciencedirect.com/science/article/pii/S0004370297000477}.
\newblock Relevance.

\bibitem[Gerber(2015)]{GERBER2015798}
Mathieu Gerber.
\newblock On integration methods based on scrambled nets of arbitrary size.
\newblock \emph{Journal of Complexity}, 31\penalty0 (6):\penalty0 798--816, 2015.
\newblock ISSN 0885-064X.
\newblock \doi{https://doi.org/10.1016/j.jco.2015.06.001}.
\newblock URL \url{https://www.sciencedirect.com/science/article/pii/S0885064X1500059X}.

\bibitem[Goyal et~al.(2019{\natexlab{a}})Goyal, Shalit, and Kim]{DBLP:journals/corr/abs-1907-07165}
Yash Goyal, Uri Shalit, and Been Kim.
\newblock Explaining classifiers with causal concept effect (cace).
\newblock \emph{CoRR}, abs/1907.07165, 2019{\natexlab{a}}.
\newblock URL \url{http://arxiv.org/abs/1907.07165}.

\bibitem[Goyal et~al.(2019{\natexlab{b}})Goyal, Wu, Ernst, Batra, Parikh, and Lee]{pmlr-v97-goyal19a}
Yash Goyal, Ziyan Wu, Jan Ernst, Dhruv Batra, Devi Parikh, and Stefan Lee.
\newblock Counterfactual visual explanations.
\newblock In Kamalika Chaudhuri and Ruslan Salakhutdinov, editors, \emph{Proceedings of the 36th International Conference on Machine Learning}, volume~97 of \emph{Proceedings of Machine Learning Research}, pages 2376--2384. PMLR, 09--15 Jun 2019{\natexlab{b}}.
\newblock URL \url{https://proceedings.mlr.press/v97/goyal19a.html}.

\bibitem[Heskes et~al.(2020)Heskes, Sijben, Bucur, and Claassen]{Causal_Shapley_value}
Tom Heskes, Evi Sijben, Ioan~Gabriel Bucur, and Tom Claassen.
\newblock Causal shapley values: Exploiting causal knowledge to explain individual predictions of complex models.
\newblock In H.~Larochelle, M.~Ranzato, R.~Hadsell, M.F. Balcan, and H.~Lin, editors, \emph{Advances in Neural Information Processing Systems}, volume~33, pages 4778--4789. Curran Associates, Inc., 2020.
\newblock URL \url{https://proceedings.neurips.cc/paper_files/paper/2020/file/32e54441e6382a7fbacbbbaf3c450059-Paper.pdf}.

\bibitem[Hooker(2007)]{eb4f7804-e31a-3e9c-abf5-60bb5e0ffa06}
Giles Hooker.
\newblock Generalized functional anova diagnostics for high-dimensional functions of dependent variables.
\newblock \emph{Journal of Computational and Graphical Statistics}, 16\penalty0 (3):\penalty0 709--732, 2007.
\newblock ISSN 10618600.
\newblock URL \url{http://www.jstor.org/stable/27594267}.

\bibitem[Ibrahim et~al.(2019)Ibrahim, Louie, Modarres, and Paisley]{GAM}
Mark Ibrahim, Melissa Louie, Ceena Modarres, and John Paisley.
\newblock Global explanations of neural networks: Mapping the landscape of predictions.
\newblock In \emph{Proceedings of the 2019 AAAI/ACM Conference on AI, Ethics, and Society}, AIES '19, page 279–287, New York, NY, USA, 2019. Association for Computing Machinery.
\newblock ISBN 9781450363242.
\newblock \doi{10.1145/3306618.3314230}.
\newblock URL \url{https://doi.org/10.1145/3306618.3314230}.

\bibitem[Irons et~al.(2021)Irons, Scetbon, Pal, and Harchaoui]{irons2021triangularflowsgenerativemodeling}
Nicholas~J. Irons, Meyer Scetbon, Soumik Pal, and Zaid Harchaoui.
\newblock Triangular flows for generative modeling: Statistical consistency, smoothness classes, and fast rates, 2021.
\newblock URL \url{https://arxiv.org/abs/2112.15595}.

\bibitem[Jansen(1999)]{JANSEN199935}
Michiel~J.W. Jansen.
\newblock Analysis of variance designs for model output.
\newblock \emph{Computer Physics Communications}, 117\penalty0 (1):\penalty0 35--43, 1999.
\newblock ISSN 0010-4655.
\newblock \doi{https://doi.org/10.1016/S0010-4655(98)00154-4}.
\newblock URL \url{https://www.sciencedirect.com/science/article/pii/S0010465598001544}.

\bibitem[Janzing et~al.(2020)Janzing, Minorics, and Bloebaum]{pmlr-v108-janzing20a}
Dominik Janzing, Lenon Minorics, and Patrick Bloebaum.
\newblock Feature relevance quantification in explainable ai: A causal problem.
\newblock In Silvia Chiappa and Roberto Calandra, editors, \emph{Proceedings of the Twenty Third International Conference on Artificial Intelligence and Statistics}, volume 108 of \emph{Proceedings of Machine Learning Research}, pages 2907--2916. PMLR, 26--28 Aug 2020.
\newblock URL \url{https://proceedings.mlr.press/v108/janzing20a.html}.

\bibitem[Janzing et~al.(2024)Janzing, Blöbaum, Mastakouri, Faller, Minorics, and Budhathoki]{janzing2024quantifyingintrinsiccausalcontributions}
Dominik Janzing, Patrick Blöbaum, Atalanti~A. Mastakouri, Philipp~M. Faller, Lenon Minorics, and Kailash Budhathoki.
\newblock Quantifying intrinsic causal contributions via structure preserving interventions, 2024.
\newblock URL \url{https://arxiv.org/abs/2007.00714}.

\bibitem[Javaloy et~al.(2023)Javaloy, Martin, and Valera]{javaloy2023causal}
Adri{\'a}n Javaloy, Pablo~Sanchez Martin, and Isabel Valera.
\newblock Causal normalizing flows: from theory to practice.
\newblock In \emph{Thirty-seventh Conference on Neural Information Processing Systems}, 2023.
\newblock URL \url{https://openreview.net/forum?id=QIFoCI7ca1}.

\bibitem[Jung et~al.(2022)Jung, Kasiviswanathan, Tian, Janzing, Bloebaum, and Bareinboim]{do-shap}
Yonghan Jung, Shiva Kasiviswanathan, Jin Tian, Dominik Janzing, Patrick Bloebaum, and Elias Bareinboim.
\newblock On measuring causal contributions via do-interventions.
\newblock In Kamalika Chaudhuri, Stefanie Jegelka, Le~Song, Csaba Szepesvari, Gang Niu, and Sivan Sabato, editors, \emph{Proceedings of the 39th International Conference on Machine Learning}, volume 162 of \emph{Proceedings of Machine Learning Research}, pages 10476--10501. PMLR, 17--23 Jul 2022.
\newblock URL \url{https://proceedings.mlr.press/v162/jung22a.html}.

\bibitem[Kancheti et~al.(2021)Kancheti, Reddy, Balasubramanian, and Sharma]{Kancheti2021MatchingLC}
Sai~Srinivas Kancheti, Abbavaram~Gowtham Reddy, Vineeth~N. Balasubramanian, and Amit Sharma.
\newblock Matching learned causal effects of neural networks with domain priors.
\newblock In \emph{International Conference on Machine Learning}, 2021.
\newblock URL \url{https://api.semanticscholar.org/CorpusID:246473224}.

\bibitem[Khemakhem et~al.(2019)Khemakhem, Kingma, and Hyv{\"a}rinen]{Khemakhem2019VariationalAA}
Ilyes Khemakhem, Diederik~P. Kingma, and Aapo Hyv{\"a}rinen.
\newblock Variational autoencoders and nonlinear ica: A unifying framework.
\newblock In \emph{International Conference on Artificial Intelligence and Statistics}, 2019.
\newblock URL \url{https://api.semanticscholar.org/CorpusID:195874364}.

\bibitem[Knuth and Szwarcfiter(1974)]{KNUTH1974153}
Donald~E. Knuth and Jayme~L. Szwarcfiter.
\newblock A structured program to generate all topological sorting arrangements.
\newblock \emph{Information Processing Letters}, 2\penalty0 (6):\penalty0 153--157, 1974.
\newblock ISSN 0020-0190.
\newblock \doi{https://doi.org/10.1016/0020-0190(74)90001-5}.
\newblock URL \url{https://www.sciencedirect.com/science/article/pii/0020019074900015}.

\bibitem[Kommiya~Mothilal et~al.(2021)Kommiya~Mothilal, Mahajan, Tan, and Sharma]{10.1145/3461702.3462597}
Ramaravind Kommiya~Mothilal, Divyat Mahajan, Chenhao Tan, and Amit Sharma.
\newblock Towards unifying feature attribution and counterfactual explanations: Different means to the same end.
\newblock In \emph{Proceedings of the 2021 AAAI/ACM Conference on AI, Ethics, and Society}, AIES '21, page 652–663, New York, NY, USA, 2021. Association for Computing Machinery.
\newblock ISBN 9781450384735.
\newblock \doi{10.1145/3461702.3462597}.
\newblock URL \url{https://doi.org/10.1145/3461702.3462597}.

\bibitem[Kucherenko et~al.(2012)Kucherenko, Tarantola, and Annoni]{KUCHERENKO2012937}
S.~Kucherenko, S.~Tarantola, and P.~Annoni.
\newblock Estimation of global sensitivity indices for models with dependent variables.
\newblock \emph{Computer Physics Communications}, 183\penalty0 (4):\penalty0 937--946, 2012.
\newblock ISSN 0010-4655.
\newblock \doi{https://doi.org/10.1016/j.cpc.2011.12.020}.
\newblock URL \url{https://www.sciencedirect.com/science/article/pii/S0010465511004085}.

\bibitem[K{\"u}gelgen et~al.(2023)K{\"u}gelgen, Mohamed, and Beckers]{kugelgen2023backtracking}
Julius~Von K{\"u}gelgen, Abdirisak Mohamed, and Sander Beckers.
\newblock Backtracking counterfactuals.
\newblock In \emph{2nd Conference on Causal Learning and Reasoning}, 2023.
\newblock URL \url{https://openreview.net/forum?id=stVikewRRvw}.

\bibitem[Kuhnt and Kalka(2022)]{Kuhnt2022}
Sonja Kuhnt and Arkadius Kalka.
\newblock \emph{Global Sensitivity Analysis for the Interpretation of Machine Learning Algorithms}, pages 155--169.
\newblock Springer International Publishing, Cham, 2022.
\newblock \doi{10.1007/978-3-031-07155-3_6}.
\newblock URL \url{https://doi.org/10.1007/978-3-031-07155-3_6}.

\bibitem[Lakkaraju et~al.(2016)Lakkaraju, Bach, and Leskovec]{Lakkaraju2016InterpretableDS}
Himabindu Lakkaraju, Stephen~H. Bach, and Jure Leskovec.
\newblock Interpretable decision sets: A joint framework for description and prediction.
\newblock \emph{Proceedings of the 22nd ACM SIGKDD International Conference on Knowledge Discovery and Data Mining}, 2016.
\newblock URL \url{https://api.semanticscholar.org/CorpusID:12533380}.

\bibitem[Larson et~al.(2016)Larson, Mattu, Kirchner, and Angwin]{compas_data}
Jeff Larson, Surya Mattu, Lauren Kirchner, and Julia Angwin.
\newblock How we analyzed the compas recidivism algorithm, 2016.
\newblock URL \url{https://www.propublica.org/article/how-we-analyzed-the-compas-recidivism-algorithm}.

\bibitem[L'Ecuyer(2018)]{RQMC_pierre}
Pierre L'Ecuyer.
\newblock Randomized quasi-monte carlo: An introduction for practitioners.
\newblock In Art~B. Owen and Peter~W. Glynn, editors, \emph{Monte Carlo and Quasi-Monte Carlo Methods}, pages 29--52, Cham, 2018. Springer International Publishing.

\bibitem[L'Ecuyer and Lemieux(2002)]{L’Ecuyer2002}
Pierre L'Ecuyer and Christiane Lemieux.
\newblock \emph{Recent Advances in Randomized Quasi-Monte Carlo Methods}, pages 419--474.
\newblock Springer US, New York, NY, 2002.
\newblock ISBN 978-0-306-48102-4.
\newblock \doi{10.1007/0-306-48102-2_20}.
\newblock URL \url{https://doi.org/10.1007/0-306-48102-2_20}.

\bibitem[Li et~al.(2010)Li, Rabitz, Yelvington, Oluwole, Bacon, Kolb, and Schoendorf]{LI}
Genyuan Li, Herschel Rabitz, Paul~E. Yelvington, Oluwayemisi~O. Oluwole, Fred Bacon, Charles~E. Kolb, and Jacqueline Schoendorf.
\newblock Global sensitivity analysis for systems with independent and/or correlated inputs.
\newblock \emph{The Journal of Physical Chemistry A}, May 2010.

\bibitem[Lundberg and Lee(2017)]{SHAP}
Scott~M. Lundberg and Su-In Lee.
\newblock A unified approach to interpreting model predictions.
\newblock In \emph{Proceedings of the 31st International Conference on Neural Information Processing Systems}, NIPS'17, page 4768–4777, Red Hook, NY, USA, 2017. Curran Associates Inc.
\newblock ISBN 9781510860964.

\bibitem[Mahajan et~al.(2019)Mahajan, Tan, and Sharma]{Mahajan2019PreservingCC}
Divyat Mahajan, Chenhao Tan, and Amit Sharma.
\newblock Preserving causal constraints in counterfactual explanations for machine learning classifiers.
\newblock \emph{ArXiv}, abs/1912.03277, 2019.
\newblock URL \url{https://api.semanticscholar.org/CorpusID:208857863}.

\bibitem[Mitchell et~al.(2022)Mitchell, Cooper, Frank, and Holmes]{10.5555/3586589.3586632}
Rory Mitchell, Joshua Cooper, Eibe Frank, and Geoffrey Holmes.
\newblock Sampling permutations for shapley value estimation.
\newblock \emph{J. Mach. Learn. Res.}, 23\penalty0 (1), January 2022.
\newblock ISSN 1532-4435.

\bibitem[Owen(1998)]{Scrambling_Sobol}
Art~B. Owen.
\newblock Scrambling sobol' and niederreiter–xing points.
\newblock \emph{Journal of Complexity}, 14\penalty0 (4):\penalty0 466--489, 1998.
\newblock ISSN 0885-064X.
\newblock \doi{https://doi.org/10.1006/jcom.1998.0487}.
\newblock URL \url{https://www.sciencedirect.com/science/article/pii/S0885064X98904873}.

\bibitem[Owen(2014)]{sobol_shapley}
Art~B. Owen.
\newblock Sobol' indices and shapley value.
\newblock \emph{SIAM/ASA Journal on Uncertainty Quantification}, 2\penalty0 (1):\penalty0 245--251, 2014.
\newblock \doi{10.1137/130936233}.
\newblock URL \url{https://doi.org/10.1137/130936233}.

\bibitem[Owen and Rudolf(2021)]{LLN_scrambled}
Art~B. Owen and Daniel Rudolf.
\newblock A strong law of large numbers for scrambled net integration.
\newblock \emph{SIAM Review}, 63\penalty0 (2):\penalty0 360--372, 2021.
\newblock \doi{10.1137/20M1320535}.
\newblock URL \url{https://doi.org/10.1137/20M1320535}.

\bibitem[Papamakarios et~al.(2017)Papamakarios, Pavlakou, and Murray]{NIPS2017_6c1da886}
George Papamakarios, Theo Pavlakou, and Iain Murray.
\newblock Masked autoregressive flow for density estimation.
\newblock In I.~Guyon, U.~Von Luxburg, S.~Bengio, H.~Wallach, R.~Fergus, S.~Vishwanathan, and R.~Garnett, editors, \emph{Advances in Neural Information Processing Systems}, volume~30. Curran Associates, Inc., 2017.
\newblock URL \url{https://proceedings.neurips.cc/paper_files/paper/2017/file/6c1da886822c67822bcf3679d04369fa-Paper.pdf}.

\bibitem[Papamakarios et~al.(2021)Papamakarios, Nalisnick, Rezende, Mohamed, and Lakshminarayanan]{NF}
George Papamakarios, Eric Nalisnick, Danilo~Jimenez Rezende, Shakir Mohamed, and Balaji Lakshminarayanan.
\newblock Normalizing flows for probabilistic modeling and inference.
\newblock \emph{J. Mach. Learn. Res.}, 22\penalty0 (1), January 2021.
\newblock ISSN 1532-4435.

\bibitem[Pawlowski et~al.(2020)Pawlowski, Castro, and Glocker]{pawlowski2020dscm}
Nick Pawlowski, Daniel~C. Castro, and Ben Glocker.
\newblock Deep structural causal models for tractable counterfactual inference.
\newblock In \emph{Advances in Neural Information Processing Systems}, 2020.

\bibitem[Pearl(2009)]{Pearl_2009}
Judea Pearl.
\newblock \emph{Causality}.
\newblock Cambridge University Press, 2 edition, 2009.

\bibitem[Petsiuk et~al.(2018)Petsiuk, Das, and Saenko]{DBLP:conf/bmvc/PetsiukDS18}
Vitali Petsiuk, Abir Das, and Kate Saenko.
\newblock {RISE:} randomized input sampling for explanation of black-box models.
\newblock In \emph{British Machine Vision Conference 2018, {BMVC} 2018, Newcastle, UK, September 3-6, 2018}, page 151. {BMVA} Press, 2018.
\newblock URL \url{http://bmvc2018.org/contents/papers/1064.pdf}.

\bibitem[Pitis et~al.(2020)Pitis, Creager, and Garg]{NEURIPS2020_294e09f2}
Silviu Pitis, Elliot Creager, and Animesh Garg.
\newblock Counterfactual data augmentation using locally factored dynamics.
\newblock In H.~Larochelle, M.~Ranzato, R.~Hadsell, M.F. Balcan, and H.~Lin, editors, \emph{Advances in Neural Information Processing Systems}, volume~33, pages 3976--3990. Curran Associates, Inc., 2020.
\newblock URL \url{https://proceedings.neurips.cc/paper_files/paper/2020/file/294e09f267683c7ddc6cc5134a7e68a8-Paper.pdf}.

\bibitem[Plečko and Bareinboim(2024)]{MAL-106}
Drago Plečko and Elias Bareinboim.
\newblock Causal fairness analysis: A causal toolkit for fair machine learning.
\newblock \emph{Foundations and Trends® in Machine Learning}, 17\penalty0 (3):\penalty0 304--589, 2024.
\newblock ISSN 1935-8237.
\newblock \doi{10.1561/2200000106}.
\newblock URL \url{http://dx.doi.org/10.1561/2200000106}.

\bibitem[Puy et~al.(2022)Puy, Becker, Piano, and Saltelli]{Puy_2022}
Arnald Puy, William Becker, Samuele~Lo Piano, and Andrea Saltelli.
\newblock A comprehensive comparison of total-order estimators for global sensitivity analysis.
\newblock \emph{International Journal for Uncertainty Quantification}, 12\penalty0 (2):\penalty0 1--18, 2022.
\newblock ISSN 2152-5080.

\bibitem[Quinlan(1993)]{auto_mpg_9}
R.~Quinlan.
\newblock {Auto MPG}.
\newblock UCI Machine Learning Repository, 1993.
\newblock {DOI}: https://doi.org/10.24432/C5859H.

\bibitem[Rahman(2014)]{doi:10.1137/120904378}
Sharif Rahman.
\newblock A generalized anova dimensional decomposition for dependent probability measures.
\newblock \emph{SIAM/ASA Journal on Uncertainty Quantification}, 2\penalty0 (1):\penalty0 670--697, 2014.
\newblock \doi{10.1137/120904378}.
\newblock URL \url{https://doi.org/10.1137/120904378}.

\bibitem[Reddy et~al.(2023{\natexlab{a}})Reddy, Bachu, Pathak, Godfrey, Balasubramanian, Varshaneya, and Kar]{Reddy2023TowardsLA}
Abbaavaram~Gowtham Reddy, Saketh Bachu, Harsh~Nilesh Pathak, Ben Godfrey, Vineeth~N. Balasubramanian, V~Varshaneya, and Satya~Narayanan Kar.
\newblock Towards learning and explaining indirect causal effects in neural networks.
\newblock In \emph{AAAI Conference on Artificial Intelligence}, 2023{\natexlab{a}}.
\newblock URL \url{https://api.semanticscholar.org/CorpusID:257756923}.

\bibitem[Reddy et~al.(2023{\natexlab{b}})Reddy, Bachu, Dash, Sharma, Sharma, and Balasubramanian]{reddy2023counterfactualdataaugmentationconfounding}
Abbavaram~Gowtham Reddy, Saketh Bachu, Saloni Dash, Charchit Sharma, Amit Sharma, and Vineeth~N Balasubramanian.
\newblock On counterfactual data augmentation under confounding, 2023{\natexlab{b}}.
\newblock URL \url{https://arxiv.org/abs/2305.18183}.

\bibitem[Retzlaff et~al.(2024)Retzlaff, Angerschmid, Saranti, Schneeberger, Röttger, Müller, and Holzinger]{RETZLAFF2024101243}
Carl~O. Retzlaff, Alessa Angerschmid, Anna Saranti, David Schneeberger, Richard Röttger, Heimo Müller, and Andreas Holzinger.
\newblock Post-hoc vs ante-hoc explanations: xai design guidelines for data scientists.
\newblock \emph{Cognitive Systems Research}, 86:\penalty0 101243, 2024.
\newblock ISSN 1389-0417.
\newblock \doi{https://doi.org/10.1016/j.cogsys.2024.101243}.
\newblock URL \url{https://www.sciencedirect.com/science/article/pii/S1389041724000378}.

\bibitem[Ribeiro et~al.(2016{\natexlab{a}})Ribeiro, Singh, and Guestrin]{LIME}
Marco~Tulio Ribeiro, Sameer Singh, and Carlos Guestrin.
\newblock "why should i trust you?": Explaining the predictions of any classifier.
\newblock In \emph{Proceedings of the 22nd ACM SIGKDD International Conference on Knowledge Discovery and Data Mining}, KDD '16, page 1135–1144, New York, NY, USA, 2016{\natexlab{a}}. Association for Computing Machinery.
\newblock ISBN 9781450342322.
\newblock \doi{10.1145/2939672.2939778}.
\newblock URL \url{https://doi.org/10.1145/2939672.2939778}.

\bibitem[Ribeiro et~al.(2016{\natexlab{b}})Ribeiro, Singh, and Guestrin]{SP_LIME}
Marco~Tulio Ribeiro, Sameer Singh, and Carlos Guestrin.
\newblock "why should i trust you?": Explaining the predictions of any classifier.
\newblock In \emph{Proceedings of the 22nd ACM SIGKDD International Conference on Knowledge Discovery and Data Mining}, KDD '16, page 1135–1144, New York, NY, USA, 2016{\natexlab{b}}. Association for Computing Machinery.
\newblock ISBN 9781450342322.
\newblock \doi{10.1145/2939672.2939778}.
\newblock URL \url{https://doi.org/10.1145/2939672.2939778}.

\bibitem[Saha and Garain(2022)]{saha2022on}
Saptarshi Saha and Utpal Garain.
\newblock On noise abduction for answering counterfactual queries: A practical outlook.
\newblock \emph{Transactions on Machine Learning Research}, 2022.
\newblock ISSN 2835-8856.
\newblock URL \url{https://openreview.net/forum?id=4FU8Jz1Oyj}.

\bibitem[Sale et~al.(2023)Sale, Hofman, Wimmer, H{\"u}llermeier, and Nagler]{sale2023second}
Yusuf Sale, Paul Hofman, Lisa Wimmer, Eyke H{\"u}llermeier, and Thomas Nagler.
\newblock Second-order uncertainty quantification: Variance-based measures.
\newblock \emph{arXiv preprint arXiv:2401.00276}, 2023.

\bibitem[Saltelli et~al.(2008)Saltelli, Marco, Terry, Francesca, Jessica, Debora, Michaela, and Stefano]{Saltelli2008GlobalSA}
Andrea Saltelli, Ratto Marco, A~Terry, Campolongo Francesca, Cariboni Jessica, Gatelli Debora, Saisana Michaela, and Tarantola Stefano.
\newblock Global sensitivity analysis: The primer.
\newblock 2008.
\newblock URL \url{https://api.semanticscholar.org/CorpusID:115957810}.

\bibitem[Scholbeck et~al.(2024)Scholbeck, Moosbauer, Casalicchio, Gupta, Bischl, and Heumann]{scholbeck2024positionpaperbridginggap}
Christian~A. Scholbeck, Julia Moosbauer, Giuseppe Casalicchio, Hoshin Gupta, Bernd Bischl, and Christian Heumann.
\newblock Position paper: Bridging the gap between machine learning and sensitivity analysis, 2024.
\newblock URL \url{https://arxiv.org/abs/2312.13234}.

\bibitem[Schwab and Karlen(2019)]{10.5555/3454287.3455204}
Patrick Schwab and Walter Karlen.
\newblock \emph{CXPlain: causal explanations for model interpretation under uncertainty}.
\newblock Curran Associates Inc., Red Hook, NY, USA, 2019.

\bibitem[Schölkopf et~al.(2021)Schölkopf, Locatello, Bauer, Ke, Kalchbrenner, Goyal, and Bengio]{9363924}
Bernhard Schölkopf, Francesco Locatello, Stefan Bauer, Nan~Rosemary Ke, Nal Kalchbrenner, Anirudh Goyal, and Yoshua Bengio.
\newblock Toward causal representation learning.
\newblock \emph{Proceedings of the IEEE}, 109\penalty0 (5):\penalty0 612--634, 2021.
\newblock \doi{10.1109/JPROC.2021.3058954}.

\bibitem[Selvaraju et~al.(2017)Selvaraju, Cogswell, Das, Vedantam, Parikh, and Batra]{GradCam}
Ramprasaath~R. Selvaraju, Michael Cogswell, Abhishek Das, Ramakrishna Vedantam, Devi Parikh, and Dhruv Batra.
\newblock Grad-cam: Visual explanations from deep networks via gradient-based localization.
\newblock In \emph{2017 IEEE International Conference on Computer Vision (ICCV)}, pages 618--626, 2017.
\newblock \doi{10.1109/ICCV.2017.74}.

\bibitem[Shapley(1953)]{Shapley+1953+307+318}
L.~S. Shapley.
\newblock \emph{17. A Value for n-Person Games}, pages 307--318.
\newblock Princeton University Press, Princeton, 1953.
\newblock ISBN 9781400881970.
\newblock \doi{doi:10.1515/9781400881970-018}.
\newblock URL \url{https://doi.org/10.1515/9781400881970-018}.

\bibitem[Shen et~al.(2022)Shen, Liu, Dong, Lian, Chen, and Zhang]{JMLR:v23:21-0080}
Xinwei Shen, Furui Liu, Hanze Dong, Qing Lian, Zhitang Chen, and Tong Zhang.
\newblock Weakly supervised disentangled generative causal representation learning.
\newblock \emph{Journal of Machine Learning Research}, 23\penalty0 (241):\penalty0 1--55, 2022.
\newblock URL \url{http://jmlr.org/papers/v23/21-0080.html}.

\bibitem[Shrikumar et~al.(2017)Shrikumar, Greenside, and Kundaje]{DeepLIFT}
Avanti Shrikumar, Peyton Greenside, and Anshul Kundaje.
\newblock Learning important features through propagating activation differences.
\newblock In \emph{Proceedings of the 34th International Conference on Machine Learning - Volume 70}, ICML'17, page 3145–3153. JMLR.org, 2017.

\bibitem[Si et~al.(2023)Si, Chen, Sahoo, Schiff, and Kuleshov]{SAFE}
Phillip Si, Zeyi Chen, Subham~Sekhar Sahoo, Yair Schiff, and Volodymyr Kuleshov.
\newblock Semi-autoregressive energy flows: exploring likelihood-free training of normalizing flows.
\newblock In \emph{Proceedings of the 40th International Conference on Machine Learning}, ICML'23. JMLR.org, 2023.

\bibitem[Simonyan et~al.(2014)Simonyan, Vedaldi, and Zisserman]{deepinsidecnn}
Karen Simonyan, Andrea Vedaldi, and Andrew Zisserman.
\newblock Deep inside convolutional networks: Visualising image classification models and saliency maps, 2014.
\newblock URL \url{https://arxiv.org/abs/1312.6034}.

\bibitem[Sobol'(1967)]{SOBOL196786}
I.M Sobol'.
\newblock On the distribution of points in a cube and the approximate evaluation of integrals.
\newblock \emph{USSR Computational Mathematics and Mathematical Physics}, 7\penalty0 (4):\penalty0 86--112, 1967.
\newblock ISSN 0041-5553.
\newblock \doi{https://doi.org/10.1016/0041-5553(67)90144-9}.
\newblock URL \url{https://www.sciencedirect.com/science/article/pii/0041555367901449}.

\bibitem[Sobol'(2001)]{SOBOL}
I.M Sobol'.
\newblock Global sensitivity indices for nonlinear mathematical models and their monte carlo estimates.
\newblock \emph{Mathematics and Computers in Simulation}, 55\penalty0 (1):\penalty0 271--280, 2001.
\newblock ISSN 0378-4754.
\newblock \doi{https://doi.org/10.1016/S0378-4754(00)00270-6}.
\newblock URL \url{https://www.sciencedirect.com/science/article/pii/S0378475400002706}.
\newblock The Second IMACS Seminar on Monte Carlo Methods.

\bibitem[Stein et~al.(2022)Stein, Raponi, Sadeghi, Bouman, Van~Ham, and Bäck]{9903639}
Bas~Van Stein, Elena Raponi, Zahra Sadeghi, Niek Bouman, Roeland C. H.~J. Van~Ham, and Thomas Bäck.
\newblock A comparison of global sensitivity analysis methods for explainable ai with an application in genomic prediction.
\newblock \emph{IEEE Access}, 10:\penalty0 103364--103381, 2022.
\newblock \doi{10.1109/ACCESS.2022.3210175}.

\bibitem[Strobl et~al.(2008)Strobl, Boulesteix, Kneib, Augustin, and Zeileis]{Strobl2008-ey}
Carolin Strobl, Anne-Laure Boulesteix, Thomas Kneib, Thomas Augustin, and Achim Zeileis.
\newblock Conditional variable importance for random forests.
\newblock \emph{BMC Bioinformatics}, 9\penalty0 (1):\penalty0 307, July 2008.

\bibitem[Sundararajan et~al.(2017)Sundararajan, Taly, and Yan]{IG}
Mukund Sundararajan, Ankur Taly, and Qiqi Yan.
\newblock Axiomatic attribution for deep networks.
\newblock In \emph{International Conference on Machine Learning}, 2017.
\newblock URL \url{https://api.semanticscholar.org/CorpusID:16747630}.

\bibitem[Tunkiel et~al.(2020)Tunkiel, Sui, and Wiktorski]{TUNKIEL2020107630}
Andrzej~T. Tunkiel, Dan Sui, and Tomasz Wiktorski.
\newblock Data-driven sensitivity analysis of complex machine learning models: A case study of directional drilling.
\newblock \emph{Journal of Petroleum Science and Engineering}, 195:\penalty0 107630, 2020.
\newblock ISSN 0920-4105.
\newblock \doi{https://doi.org/10.1016/j.petrol.2020.107630}.
\newblock URL \url{https://www.sciencedirect.com/science/article/pii/S0920410520306975}.

\bibitem[Van~Looveren and Klaise(2021)]{10.1007/978-3-030-86520-7_40}
Arnaud Van~Looveren and Janis Klaise.
\newblock Interpretable counterfactual explanations guided by prototypes.
\newblock In Nuria Oliver, Fernando P{\'e}rez-Cruz, Stefan Kramer, Jesse Read, and Jose~A. Lozano, editors, \emph{Machine Learning and Knowledge Discovery in Databases. Research Track}, pages 650--665, Cham, 2021. Springer International Publishing.
\newblock ISBN 978-3-030-86520-7.

\bibitem[Verma et~al.(2022)Verma, Boonsanong, Hoang, Hines, Dickerson, and Shah]{verma2022counterfactualexplanationsalgorithmicrecourses}
Sahil Verma, Varich Boonsanong, Minh Hoang, Keegan~E. Hines, John~P. Dickerson, and Chirag Shah.
\newblock Counterfactual explanations and algorithmic recourses for machine learning: A review, 2022.
\newblock URL \url{https://arxiv.org/abs/2010.10596}.

\bibitem[Wachter et~al.(2018)Wachter, Mittelstadt, and Russell]{wachter2018counterfactualexplanationsopeningblack}
Sandra Wachter, Brent Mittelstadt, and Chris Russell.
\newblock Counterfactual explanations without opening the black box: Automated decisions and the gdpr, 2018.
\newblock URL \url{https://arxiv.org/abs/1711.00399}.

\bibitem[Wang et~al.(2023)Wang, Liu, Chen, Wu, Hao, Chen, and Heng]{9982682}
Yunqi Wang, Furui Liu, Zhitang Chen, Yik-Chung Wu, Jianye Hao, Guangyong Chen, and Pheng-Ann Heng.
\newblock Contrastive-ace: Domain generalization through alignment of causal mechanisms.
\newblock \emph{IEEE Transactions on Image Processing}, 32:\penalty0 235--250, 2023.
\newblock \doi{10.1109/TIP.2022.3227457}.

\bibitem[Wimmer et~al.(2023)Wimmer, Sale, Hofman, Bischl, and H\"ullermeier]{pmlr-v216-wimmer23a}
Lisa Wimmer, Yusuf Sale, Paul Hofman, Bernd Bischl, and Eyke H\"ullermeier.
\newblock Quantifying aleatoric and epistemic uncertainty in machine learning: Are conditional entropy and mutual information appropriate measures?
\newblock In Robin~J. Evans and Ilya Shpitser, editors, \emph{Proceedings of the Thirty-Ninth Conference on Uncertainty in Artificial Intelligence}, volume 216 of \emph{Proceedings of Machine Learning Research}, pages 2282--2292. PMLR, 31 Jul--04 Aug 2023.
\newblock URL \url{https://proceedings.mlr.press/v216/wimmer23a.html}.

\bibitem[Xi and Bloem{-}Reddy(2023)]{DBLP:conf/aistats/XiB23}
Quanhan Xi and Benjamin Bloem{-}Reddy.
\newblock Indeterminacy in generative models: Characterization and strong identifiability.
\newblock In Francisco J.~R. Ruiz, Jennifer~G. Dy, and Jan{-}Willem van~de Meent, editors, \emph{International Conference on Artificial Intelligence and Statistics, 25-27 April 2023, Palau de Congressos, Valencia, Spain}, volume 206 of \emph{Proceedings of Machine Learning Research}, pages 6912--6939. {PMLR}, 2023.
\newblock URL \url{https://proceedings.mlr.press/v206/xi23a.html}.

\bibitem[Yadu et~al.(2021)Yadu, Suhas, and Sinha]{9506118}
Ankit Yadu, P~K Suhas, and Neelam Sinha.
\newblock Class specific interpretability in cnn using causal analysis.
\newblock In \emph{2021 IEEE International Conference on Image Processing (ICIP)}, pages 3702--3706, 2021.
\newblock \doi{10.1109/ICIP42928.2021.9506118}.

\bibitem[Yang et~al.(2018)Yang, Rangarajan, and Ranka]{8622994}
Chengliang Yang, Anand Rangarajan, and Sanjay Ranka.
\newblock Global model interpretation via recursive partitioning.
\newblock In \emph{2018 IEEE 20th International Conference on High Performance Computing and Communications; IEEE 16th International Conference on Smart City; IEEE 4th International Conference on Data Science and Systems (HPCC/SmartCity/DSS)}, pages 1563--1570, 2018.
\newblock \doi{10.1109/HPCC/SmartCity/DSS.2018.00256}.

\bibitem[Yang et~al.(2021)Yang, Liu, Chen, Shen, Hao, and Wang]{9578520}
Mengyue Yang, Furui Liu, Zhitang Chen, Xinwei Shen, Jianye Hao, and Jun Wang.
\newblock Causalvae: Disentangled representation learning via neural structural causal models.
\newblock In \emph{2021 IEEE/CVF Conference on Computer Vision and Pattern Recognition (CVPR)}, pages 9588--9597, 2021.
\newblock \doi{10.1109/CVPR46437.2021.00947}.

\bibitem[Zeiler and Fergus(2014)]{Deconvolutional}
Matthew~D. Zeiler and Rob Fergus.
\newblock Visualizing and understanding convolutional networks.
\newblock In David Fleet, Tomas Pajdla, Bernt Schiele, and Tinne Tuytelaars, editors, \emph{Computer Vision -- ECCV 2014}, pages 818--833, Cham, 2014. Springer International Publishing.
\newblock ISBN 978-3-319-10590-1.

\bibitem[Zhu et~al.(2020)Zhu, Ng, and Chen]{Zhu2020Causal}
Shengyu Zhu, Ignavier Ng, and Zhitang Chen.
\newblock Causal discovery with reinforcement learning.
\newblock In \emph{International Conference on Learning Representations}, 2020.
\newblock URL \url{https://openreview.net/forum?id=S1g2skStPB}.

\end{thebibliography}
\newpage
\appendix


\section{Structural Causal Models}\label{SCM}

\subsection{Causality Preliminaries \citep{kugelgen2023backtracking,reddy2023counterfactualdataaugmentationconfounding}}
In this Section, we outline the fundamental definitions and concepts necessary to understand our paper. 
\begin{definition}[Structural Causal Models]
A Structural Causal Model (SCM) $\mathcal{S}(\mathbf{V},\mathbf{U},\mathbf{f}, P_{\mathbf{U}})$ represents cause-effect relationships among a set of random variables, divided into endogenous variables $\mathbf{V}=\{V_{1},V_{2},...,V_{p} \}$ and exogenous variables $\mathbf{U}=\{U_{1}, U_{2},..., U_{p} \}$, through a collection of structural equations 
$\mathbf{f}=\{f_{1},f_{2},...,f_{p}\}$. Each variable $V_{j} \in \mathbf{V}$ is defined in relation to its parents $PA_{j} \subseteq \mathbf{V}-j$ using the causal law $V_{j}=f_{j}(PA_{j},U_{j})$. $P_{\mathbf{U}}$ is the probability distribution over exogenous variables.
\end{definition}

The causal diagram $\mathfrak{G}(\mathcal{S})$ affiliated with an SCM $\mathcal{S}$ is a directed graph where each node represents a variable, and directed edges point from the elements of $PA_{j}$ and $U_{j}$ towards $V_{j}$. As exogenous variables $
\mathbf{U}$ are typically unobserved, it is common practice to focus only on the subset of
$\mathfrak{G}(\mathcal{S})$ projected onto 
$\mathbf{V}$. 
A directed graph is acyclic if it contains no cycles, in which case it is called a directed acyclic graph (DAG).
A path in a causal graph (DAG)  is defined as a sequence of distinct vertices $X_{1}, X_{2}, ..., X_{n}$ such that there is an edge between each pair of consecutive vertices $
X_{i} $ and $X_{i+1}$. This edge can be either $X_{i}\rightarrow X_{i+1}$ or $X_{i+1}\rightarrow X_{i}$. A directed path is one where all edges point in the same direction. If there is a directed path from $X_{j}$ to $X_{i}$, then $X_{j}$ is called an ancestor of $X_{i}$, and $X_{i}$ is referred to as a descendant of $X_{j}$.

\begin{definition}[Interventional Distribution]
The
interventional distribution $\mathbf{X}$ under an intervention where 
 $X_{j}$ is set to a specific value $x_{j}$, denoted as $do(X_{j}=x_{j})$, is defined as follows: 
\begin{align*}
 P(\mathbf{X}| do(X_{j}=x_{j})) =\mathbf{1}_{X_{j}=x_{j}} \times \prod_{i\neq j} P(X_{i}|PA_{i}),
\end{align*}
where $\mathbf{1}$ is the indicator function.
\end{definition}

\section{Causal Interpretation}\label{causal_interpretation}
Using the backtracking semantics from
\cite{kugelgen2023backtracking}, we will now explain the causal interpretation of ICC. “Backtracking” alludes to the process of adjusting upstream variables to account for counterfacts while preserving the underlying causal structures in the system.


Instead of measuring the reduction in uncertainty caused by adjusting the observed value $x_{j}$ of the node $X_{j}$, we assess the reduction achieved by modifying the associated noise $u_{j}$. 
However, through the backtracking, adjustment of a noise $u_{j}$ can be interpreted as
an intervention on $X_{j}$ without altering the joint distribution of $\mathbf{X}$ \citep{janzing2024quantifyingintrinsiccausalcontributions} : 
After noting that the parent variables of 
$X_{j}$ have taken the values $pa_{j}$, we assign $X_{j}$ the value $x_{j}'= f_{j}(pa_{j},u_{j}')$, where $u_{j}'$ is randomly sampled from $P_{U_{j}}$. 
As $U_{j}$ has no parents, we can treat $U_{j}$ as a randomized treatment. Thus, the interventional probabilities are reduced to observational probabilities
\begin{align*}
 P(\cdot|do(U_{j}=u_{j}')) = P(\cdot|U_{j}=u_{j}'), \qquad \forall 1\leq j\leq p.
\end{align*}
As a result, we did not explicitly
write interventional probabilities in the Definition \ref{def_ICC}. However, a statistical dependence between $\hat{Y}$ and $U_{j}$ suggests a causal influence of $X_{j}$ on $\hat{Y}$.

\section{Variance-based Sensitivity Analysis}\label{VSA}
The Hoeffding decomposition, also known as the ANOVA decomposition or high-dimensional model representation (HDMR), allows us to represent the function 
$f$ as follows:
\begin{align}
 \hat{Y}=f(\mathbf{X}) = \sum_{T\subseteq[p]}f_{T}(\mathbf{X}_{T}),
\end{align}
with the functions $f_{T}$ defined recursively as
\begin{align}\label{anova2}
f_{T}(\mathbf{X}_{T}) = \mathbb{E}(f(\mathbf{X})|\mathbf{X}_{T}) - \sum_{T'\subset T} f_{T'}(\mathbf{X}_{T'}). 
\end{align}
The ANOVA decomposition satisfies the orthogonality constraint
\begin{align}\label{anova3}
 \mathbb{E}_{T\neq T'}(f_{T}(\mathbf{X}_{T})f_{T'}(\mathbf{X}_{T'})) =0 \quad \forall T,T' \subseteq[p], 
\end{align}
and let us decompose the model variance as follows:
\begin{align}
 \mathbb{V}(f(\mathbf{X})) = \sum_{T\subseteq[p]} \mathbb{V}(f_{T}(\mathbf{X}_{T})). 
\end{align}
\begin{definition}[Sobol' indices \citep{Look_at_variance}] The sensitivity index $\mathcal{S}_{T}$, which quantifies the contribution of the variable set $\mathbf{X}_{T}$ to the model response $f(\mathbf{X})$ in terms of its fluctuations, is defined as:
\begin{align}
 \mathcal{S}_{T} = \dfrac{\mathbb{V}(f_{T}(\mathbf{X}_{T}))}{\mathbb{V}(f(\mathbf{X}))}.
\end{align}
\end{definition}
Sobol indices quantify the proportion of the output’s variance caused by
any subset of input features. Theorem 1 from \citet{sobol_shapley} establishes a connection between Sobol’ indices and the Shapley value, where the latter is computed using a variance-based value function. We restate this theorem to match our setting.
\begin{theorem}\label{Sobol_shapley_thm}
For any $1 \leq j\leq p$, 
\begin{align}
 \sum_{T \subseteq [p]/\{ j\} } \dfrac{1}{p \binom{p-1}{|T|} } \Big( \mathbb{V}(\mathbb{E}(f(\mathbf{X})|\mathbf{X}_{T+j})) - \mathbb{V}(\mathbb{E}(f(\mathbf{X})|\mathbf{X}_{T})) \Big)= \sum_{T\subseteq[p], j\in T} \dfrac{\mathcal{S}_{T}}{|T|}. 
 \end{align}
\end{theorem}

\section{Proofs}\label{Proof}

\printProofs

\section{Algorithms}\label{algo_appendix}
\begin{algorithm}[H]
\caption{Pseudocode for estimating $\phi$ for $ICC^{\text{To}}$}
\label{estimation_phi_ICC_to}
\begin{algorithmic}[1] 
 \Statex \textbf{Input:} Batch size $B$, context $I$ for conditioning, the neural network $\mathcal{N}$

 \State $\mathbf{x}_{M}^{(i)}, \mathbf{x}_{N}^{(i)}\sim D $ for $i=1,2,...,B$, where $D$ is the dataset.
 \State $\mathbf{x}_{Q}^{(i)} = \Big( \mathbf{x}_{M_{-I}}^{(i)},\mathbf{x}_{N_{I}}^{(i)} \Big) $ for $i=1,2,...,B.$

 \State $\hat{y}_{M}^{(i)} = \mathcal{N}\Big(\mathbf{x}_{M}^{(i)}\Big), \hat{y}_{N}^{(i)} = \mathcal{N}\Big(\mathbf{x}_{N}^{(i)}\Big), \hat{y}_{Q}^{(i)} = \mathcal{N}\Big(\mathbf{x}_{Q}^{(i)}\Big)$ for $i=1,2,...,B.$ 

 \State $\Bar{y} = \dfrac{1}{2B}\sum_{i=1}^{B} \Big(\hat{y}_{M}^{(i)}+\hat{y}_{N}^{(i)}\Big)$; $V = \dfrac{1}{2B-1}\sum_{i=1}^{B}\Big((\hat{y}_{M}^{(i)}-\Bar{y})^{2}+(\hat{y}_{N}^{(i)}-\Bar{y})^{2}\Big)$
 
 \State $\hat{\psi} = V - \dfrac{1}{2B}\sum_{i=1}^{B} \big(\hat{y}_{N}^{(i)} - \hat{y}_{Q}^{(i)}\big)^{2}$
 
 \Statex \textbf{Output:} $\hat{\phi}=\dfrac{\hat{\psi}}{V}$
\end{algorithmic}
\end{algorithm}


\section{Experiment Setup and Datasets} \label{appendix_results}

We usually train the neural networks $\mathcal{N}$ and NFs for 100 epochs with a batch size of 32, using the Adam optimizer with a learning rate of $3\times 10^{-4}$. Only for COMPAS dataset, we train the neural network using a learning rate of $10^{-3}$.

\paragraph{Synthetic Data}
We employ the same data generation process as \citet{Reddy2023TowardsLA} for the synthetic data experiment. Figure \ref{causal_graphs_experiments}b contains the causal graph and the detailed specification of the SCM. In this dataset, the input features $W, Z,$ and $X$ are connected through linear equations with additive Gaussian noise. The output $Y$ is a non-linear function of these inputs, also incorporating additive Gaussian noise. The training set consists of 700 samples, while the test set contains 300 samples.


\paragraph{Auto-MPG}
 We use the Auto-MPG dataset to predict miles per gallon (MPG) based on features including the number of cylinders (C), displacement (D), weight (W), horsepower (H), acceleration (A), and miles per gallon (M). The ground truth causal graph for Auto-MPG is unknown, so we adopt the causal graph proposed by \cite{Reddy2023TowardsLA}, shown in Figure \ref{causal_graphs_experiments}c. This graph is constructed using relevant domain knowledge and validated through consultations with GPT-3.5 \citep{NEURIPS2020_1457c0d6} to confirm the correctness of each causal edge. The training set includes 274 samples, and the test set includes 118 samples.

\paragraph{COMPAS}
The dataset comprises criminal records and demographic features for 6,172 defendants who were released on bail in U.S. state courts between 1990 and 2009. 
The objective herein is to classify each defendant into one of two categories: bail (indicating they are unlikely to commit a violent crime if released) or no bail (indicating they are likely to commit a violent crime). The causal graph in Figure \ref{causal_graphs_experiments}a for the COMPAS dataset is inspired by \cite{MAL-106}. The training set comprises 4,937 samples, while the test set comprises 1,235 samples.

\section{Comparison with \citet{janzing2024quantifyingintrinsiccausalcontributions}} 
The fundamental difference between their work and ours is that their work does not aim to explain a downstream pre-trained model (neural networks, in our case), whereas this is the primary objective of our study.
While \citet{janzing2024quantifyingintrinsiccausalcontributions} introduce the notion of ICC, they do not provide a clear, general algorithm for its estimation. Instead, their experimental setup relies on restrictive assumptions --- such as treating an additive noise model  as a convenient approximation of structural equations (AutoMPG), or inferring the SCM (River flows) from common-sense knowledge with all regression coefficients set to 1. 
Moreover, they do not discuss identifiability.
In contrast, our framework, which incorporates causal normalizing flows, is more general and ensures identifiability.

\citet{janzing2024quantifyingintrinsiccausalcontributions} utilized the publicly available ICC implementation in DoWhy \citep{blobaum2024dowhy}.
Specifically, the \texttt{gcm.intrinsic\_causal\_influence} function was used with the auto-assign feature. The \texttt{gcm.intrinsic\_causal\_influence} function returns ICC within a causal model but is not designed to generate explanations for a pre-trained neural network.
For each node in the causal graph, \texttt{gcm.auto.assign\_causal\_mechanisms} fits various regression or classification models and selects the optimal one.
In contrast, we model the entire causal data-generating process of the inputs using a single deep neural network \citep{javaloy2023causal}.

\end{document}